**Mechanisms for Handling Nested Dependencies in Neural-Network Language Models and Humans**

Yair Lakretz

Cognitive Neuroimaging Unit, NeuroSpin center, 91191, Gif-sur-Yvette, France

Dieuwke Hupkes

ILLC, University of Amsterdam, Amsterdam, Netherlands

Alessandra Vergallito

University of Milano-Bicocca, Milan, Italy

Marco Marelli

University of Milano-Bicocca, Milan, Italy

NeuroMI, Milan Center for Neuroscience, Milan, Italy

Marco Baroni[*]

Facebook AI Research, Paris, France

Catalan Institute for Research and Advanced Studies, Barcelona, Spain, 08010

Departament de Traducció i Ciències del Llenguatge, Universitat Pompeu Fabra, Spain, 08018

Stanislas Dehaene[*]

Cognitive Neuroimaging Unit, NeuroSpin center, 91191, Gif-sur-Yvette, France

College de France, 11 Place Marcelin Berthelot, 75005 Paris, France

---

[*] Equal senior contribution.

**Abstract**

Recursive processing in sentence comprehension is considered a hallmark of human linguistic abilities. However, its underlying neural mechanisms remain largely unknown. We studied whether a modern artificial neural network trained with "deep learning" methods mimics a central aspect of human sentence processing, namely the storing of grammatical number and gender information in working memory and its use in long-distance agreement (e.g., capturing the correct number agreement between subject and verb when they are separated by other phrases). Although the network, a recurrent architecture with Long Short-Term Memory units, was solely trained to predict the next word in a large corpus, analysis showed the emergence of a very sparse set of specialized units that successfully handled local and long-distance syntactic agreement for grammatical number. However, the simulations also showed that this mechanism does not support full recursion and fails with some long-range embedded dependencies. We tested the model's predictions in a behavioral experiment where humans detected violations in number agreement in sentences with systematic variations in the singular/plural status of multiple nouns, with or without embedding. Human and model error patterns were remarkably similar, showing that the model echoes various effects observed in human data. However, a key difference was that, with embedded long-range dependencies, humans remained above chance level, while the model's systematic errors brought it below chance. Overall, our study shows that exploring the ways in which modern artificial neural networks process sentences leads to precise and testable hypotheses about human linguistic performance.

**Mechanisms for Handling Nested Dependencies in Neural-Network Language Models and Humans**

## Introduction

According to a popular view in linguistics, the rich expressiveness and open-ended nature of language rest on *recursion*, the possibly uniquely human ability to process nested structures (Chomsky, 1957; Dehaene, Meyniel, Wacongne, Wang, & Pallier, 2015; Hauser, Chomsky, & Fitch, 2002). We currently know very little about how such ability is implemented in the brain, and consequently about its scope and limits.

In recent years, artificial neural networks, rebranded as "deep learning" (LeCun, Bengio, & Hinton, 2015), made tremendous advances in natural language processing (Goldberg, 2017). Typically, neural networks are not provided with any explicit information regarding grammar or other types of linguistic knowledge. Neural Language Models (NLMs) are commonly initialized as "tabula rasa" and are solely trained to predict the next word in a sentence given the previous words (Mikolov, 2012). Yet, these models achieve impressive performance, not far below that of humans, in tasks such as question answering or text summarization (Radford et al., 2019).

Unlike their connectionist progenitors (Rumelhart, McClelland, & PDP Research Group, 1986a, 1986b), modern NLMs are not intended as cognitive models of human behaviour. They are instead developed to be deployed in applications such as translation systems or online content organizers. Nevertheless, *recurrent* NLMs (Elman, 1990; Hochreiter & Schmidhuber, 1997), at least, do share some relevant attributes with the human processing system, such as the property that they receive input incrementally and process it in a parallel and distributed way. Furthermore, their success in applied natural language processing strongly suggests that they must infer non-trivial linguistic knowledge from the raw data they are exposed to. Combined, these two facts make NLMs akin to an interesting 'animal model', whose way to address a linguistic task might provide insight into how the human processing system tackles similar challenges (see also McCloskey, 1991).

Here, we perform an in-depth analysis of nested sentence processing in NLMs, focusing in particular on the case of grammatical agreement. Long-distance agreement has traditionally been studied as one of the best indices of online syntactic processing in humans, as it is ruled by hierarchical structures rather than by the linear order of words in a sentence (Bock & Miller, 1991; Franck, Vigliocco, & Nicol, 2002). Consider for example the sentence: "The **boys** under the tree **know** the farmers", where the number of the verb ('know') depends on its linearly distant subject ('boys'), and not on the immediately preceding noun 'tree'.

Analogously, number agreement has also become a standard way to probe grammatical generalization in NLMs (Bernardy & Lappin, 2017; Giulianelli, Harding, Mohnert, Hupkes, & Zuidema, 2018; Gulordava, Bojanowski, Grave, Linzen, & Baroni, 2018; Linzen, Dupoux, & Goldberg, 2016). Very recently, some steps were taken towards a mechanistic understanding of how NLMs perform agreement. Specifically, Lakretz et al. (2019) showed that NLMs trained on a large corpus with English sentences developed a number-propagation mechanism for long-range dependencies. The core circuit of this mechanism is extremely sparse, in the sense that it is comprised of an exceptionally small number of units. Here, we further investigate this circuitry, test how robust this finding is across languages and grammatical features and, importantly, if this knowledge about the mechanism in recurrent NLMs can be useful to learn more about *human language processing.*

Our starting point is the sparsity of the mechanism found by Lakretz et al. (2019), in which number information is carried only in two single units (a plural and a singular one). The recursive power of language allows the construction of sentences with multiple nested agreement dependencies – as in: "The **boys** that the *father* under the tree *watches* **know** the farmers". Lakretz et al. (2019) showed that the NLM's mechanism is robust to the intervention of such nested hierarchical structures, across several different syntactic structure. In sentences like the example above, it could correctly percolate the number across the outermost long-distance dependency ('boy/knows'). However, the sparsity of the mechanism suggests that the NLM should not be able to process two long-distance dependencies simultaneously: once the mechanism is "filled" by the outermost long-distance dependency ('boy/knows') it is not anymore able to also track the number information of the inner long-distance dependency ('father/watches'). In other words, it predicts a failure of the model to handle the *inner-most* long-distance dependency in doubly nested sentences. In this paper, we test this prediction and then use our mechanistic understanding of agreement in NLMs as a "hypothesis generator" (Cichy & Kaiser, 2019) about nested agreement processing in humans.

We start our current study by confirming that the emergence of a sparse agreement mechanism is a stable and robust phenomenon in recurrent NLMs, by replicating the findings of Lakretz et al. (2019) with a new language (Italian) and grammatical feature (gender). Next, we investigate how the sparsity of the agreement mechanism affects recursive agreement processing, confirming our predictions that the mechanism supports outermost agreement across nested structures, but not the inner dependency in multiple embedded agreements.

In the next part of our study, we test if this pattern is also observed in humans: suppose that humans also use a relatively small fraction of specialized units to store and release agreement features across long-distance syntactic structures. Then, we might observe a similar asymmetry in handling outer- and innermost dependencies in recursive structures. We run a behavioural experiment with Italian subjects to test this hypothesis. The results are intriguing. On the one hand, humans do not display the same dramatic failure to process embedded dependencies we observed in NLMs. On the other hand, they are indeed more prone to errors in embedded dependencies than in the longer-range outer ones, in accordance with our predictions. Moreover, a comparison between NLM and human performance reveals a remarkable degree of similarity.

Our results thus indicate that NLMs do not achieve genuine recursive processing of nested long-range agreements. However, they also show how some degree of hierarchical processing can be performed by a device, such as the NLM, that did not develop full recursive capabilities. Furthermore, the similarity between the error patterns of humans and the model illustrate how a detailed understanding of emergent mechanisms in NLMs leads to hypotheses about hierarchical structure processing that are relevant to human language parsing.

# 1 Number Agreement in Neural Language Models

In the classic number agreement task (NA-task), subjects are presented with the beginning of a sentence (aka, 'preamble') that contains a long-range subject-verb relation, such as: "The **keys** to the cabinet...", and are asked to predict the verb to follow (e.g., "are"). Human subjects make more agreement errors (e.g., continuing the preamble above with "is" instead of "are") when the intervening noun (aka, 'attractor') has a different grammatical number than the main subject (as in the preamble above, with plural subject "keys" and singular attractor "cabinet") (Bock & Miller, 1991). Behavioral measures collected during agreement tasks, such as error rates, vary as a function of the syntactic environment of the long-range dependency. This has provided rich data to test hypotheses regarding online syntactic processing in humans (e.g., Franck, Frauenfelder, & Rizzi, 2007; Franck, Lassi, Frauenfelder, & Rizzi, 2006; Franck et al., 2002).

Starting with the influential work of Linzen et al. (2016), a growing literature (e.g., Bernardy & Lappin, 2017; Giulianelli et al., 2018; Gulordava et al., 2018; Jumelet, Zuidema, & Hupkes, 2019; Kuncoro et al., 2018; Linzen & Leonard, 2018) has tested NLMs on the NA-task at the behavioural level, showing that these models have performance and error patterns partially resembling those of humans. Recently, Lakretz et al. (2019) investigated the underlying mechanism of an English NLM during the processing of a long-range dependency. They identified a neural circuit in the network that encodes and carries grammatical number across long-range dependencies, showing also that processing in NLMs is sensitive to the structure of the subject-verb dependency. We now describe the main findings in this previous study, followed by a replication of the results in an NLM trained on Italian (Section 2).

## 1.1 The NounPP Number-Agreement Task

The main NA-task used by Lakretz et al. (2019) contains sentences with a subject-verb dependency separated by a prepositional phrase containing an attractor (e.g., "The **boy** near the car **smiles**"), referred to as the 'NounPP' task. This task comprises four conditions, which correspond to the four possible assignments of grammatical number to the main subject and attractor (SS, SP, PS and PP; S-singular, P-plural). A NLM whose weights were optimized on the generic objective of predicting the next word in a Wikipedia-based corpus was presented with preambles of sentences from this task. The predictions of the model for the next word were then extracted, from which error rates were computed.

## 1.2 Long-Range Number Units

Having verified that the network could predict the correct number with high accuracy, Lakretz et al. (2019) tested whether there are units in the network that are crucial to carry grammatical number across long-range dependencies. To identify such units, they conducted an ablation study, ablating one unit of the NLM at a time, and re-evaluating its performance on the NounPP task. These ablation studies showed that two (out of 1300) units in the NLM cause a reduction in long-distance agreement performance towards chance level when ablated (short-distance agreement in other NA-tasks was not affected). One of these units only affects performance when the main subject of the sentence is singular, and was therefore called the 'singular unit'. The other unit has an effect with plural subjects only, hence, the 'plural unit'. No other unit has a comparable effect on network performance when ablated. A visualization of state dynamics of the singular and plural units confirmed their role in encoding and carrying through grammatical number across long-range dependency, robustly also in the presence of an attractor (Figure 1 in Lakretz et al., 2019).

## 1.3 Syntax Units

Since the activities of the long-range number units follow the structure of the syntactic long-range dependency, Lakretz et al. (2019) tested whether other units in the network encode syntactic structure, letting the long-range number units know when to store and release number information in their encoding. Several units were found to have activity that is predictive about transient syntactic properties of the sentence. In particular, the

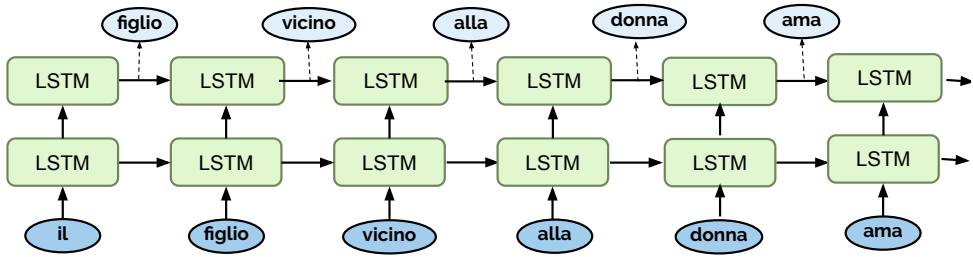

**Figure 1**

*Graphical description of a two-layer recurrent neural language model with LSTM cells (not discussed here; see, e.g., Goldberg, 2017). At each timestep, the model processes an input word and outputs a probability distribution over potential next words in the sentence. The prediction of the output word depends on both the input word and on the previous state of the model, which serves as longer-term context (horizontal arrows in the figure represent the* recurrent *connections carrying the previous state through).*

activation of one of these 'syntax' units followed the structure of the main subject-verb dependency, consistently across various sentence constructions (Figure 3 in Lakretz et al., 2019).

### 1.4 Short-Range Number Units

Lakretz et al. (2019) further found that grammatical number information is also encoded by other units in the network in a distributed way. Number information can still be decoded from network activity even when the long-range number units are removed. However, the information encoded in these other units is short-lived. Whenever a new grammatical number is introduced (e.g., upon encountering a noun or a verb), activity in these units abruptly switches to represent this last encountered number. These 'Short-Range Number Units' can therefore only support number-agreement dependencies that do not enfold attractors (e.g., "The **boy** gracefully **smiles**"). The presence of short-range number units explains why ablating the long-range circuit only affects agreement in long-distance dependencies.

## 2 Processing of a Single Subject-Verb Dependency in the Italian NLM

We focus here on Italian instead of English for three reasons. First, we aimed to replicate the English results on another language. Second, the Italian NLM made available by Gulordava et al. (2018) achieves better performance than the English NLM on nested constructions, which are computationally more demanding compared to sentences in the NounPP task explored in Lakretz et al. (2019), and which will constitute the main object of our study. This might be explained by the fact that Italian is morpho-syntactically richer than English, making it more important for the NLM to pick up grammatical information. Third, in English, the plural is identical to the unmarked form of the verb, which can occur as infinitive, in first and second person, and often as a noun. This makes the occurrence statistics extremely unbalanced in favor of plural.*

Furthermore, agreement in Italian can also occur with respect to gender, for instance, in sentences containing predicative adjectives: "il **bambino** accanto alla madre è **bello**" ("The (m.) **boy** near the (f.) mother is pretty (m.)"), in which subject and predicate agree with respect to both number and gender. This allowed us to test whether the previous results in English hold for another grammatical feature. We hypothesized that there should also exist long-range 'gender' units in the network, with dynamics that are robust to possible attractors (e.g., to "madre" above, which is feminine).

In this experiment, we followed the same steps described in the previous section for English–an ablation study, visualization of unit dynamics and a connectivity analysis.*

---

* Gulordava et al. (2018) made available models in English, Italian, Hebrew and Russian, which were optimized by conducting a grid search on the language modeling objective, and became since a subject of research in several subsequent studies (Futrell et al., 2019; Giulianelli et al., 2018; Jumelet et al., 2019; Wilcox, Levy, Morita, & Futrell, 2018). We directly use their model without any further tuning.

* Due to lack of relevant resources, we do not attempt to track down syntax units. The very fact that the Italian NLM handles various long-distance agreement tasks correctly, as shown here and in the next section, suggests that similar units must control the dynamics of its long-range number units. Proving their presence is in any case not crucial to our main predictions.

## 2.1 Materials and Methods

### 2.1.1 Agreement Tasks

For the ablation studies, we constructed two *agreement tasks* with a single long-range dependency across a prepositional phrase (Lakretz et al., 2019): NounPP-number and NounPP-gender. In the NounPP-number task, the main subject and verb agree on grammatical number, and a second noun (attractor) can interfere during sentence processing. This task was used to identify long-range number units. In the NounPP-gender task, the main subject and a predicative adjective agree on gender, and an attractor having an opposite gender can interfere in the middle. This task was used to identify long-range gender units. Table 1 (top) describes the two tasks.

### 2.1.2 Neural Language Model

In computational linguistics, a *language model* defines a probability distribution over sequences of words (Jurafsky & Martin, 2020). A Neural Language Model (NLM) is simply a language model implemented by a neural network. The recurrent NLM we use here, schematically illustrated in Figure 1, factorizes the probability of a sentence into a multiplication of the conditional probabilities of all words in the sentence, given the words that precede them:

$$P(w_1 w_2 \cdots w_n) = \prod_{i=1}^{n} p(w_i | w_1 \cdots w_{i-1}) \tag{1}$$

This type of language model can thus be used as *next-word predictor*: given the preamble of a sentence, it outputs a probability distribution over potential next words. We exploit this fact in our experiments.

### 2.1.3 Model Description

The specific NLM we used in our experiments is the Italian NLM made available by Gulordava et al. (2018).[*] It is a recurrent LSTM language model (Graves, 2012), consisting of two layers, each with 650 Long-Short Term Memory units (Hochreiter &

---

[*] https://github.com/facebookresearch/colorlessgreenRNNs

*Agreement tasks for ablation experiments*

| | | |
|---|---|---|
| *NounPP-number* | `NP`$_a$ `prep NP`$_b$ `V`$_a$ | Il **ragazzo** accanto alla <u>donna</u> **conosce** |
| | | (The **boy** next to the <u>woman</u> *knows*) |
| *NounPP-gender* | `NP`$_a$ `prep NP`$_b$ `BE`$_a$ `ADJ`$_a$ | Il **ragazzo** accanto alla <u>donna</u> è **basso** |
| | | (The **boy** next to the <u>woman</u> is **short-m**) |

*Number-agreement tasks for nesting experiments*

| | | |
|---|---|---|
| *Short-Successive* | `NP`$_a$ `V`$_a$ `che NP`$_b$ `V`$_b$ | Il **figlio dice** che il *ragazzo ama* |
| | | The **son says** that the *boy loves* |
| *Long-Successive* | `NP`$_a$ `V`$_a$ `che NP`$_b$ `P NP`$_c$ `V`$_b$ | Il **figlio dice** che l'*amico* accanto al <u>ragazzo</u> *conosce* |
| | | The **son says** that the *friend* next to the <u>boy</u> *knows* |
| *Short-Nested* | `NP`$_a$ `che NP`$_b$ `V`$_b$ `V`$_a$ | Il **figlio** che il *ragazzo osserva* **evita** |
| | | The **son** that the *boy observes* **avoids** |
| *Long-Nested* | `NP`$_a$ `che NP`$_b$ `P NP`$_c$ `V`$_b$ `V`$_a$ | Il **figlio** che la *ragazza* accanto ai <u>padri</u> *ama* **evita** |
| | | The **son** that the *girl* next to the <u>fathers</u> *loves* **avoids** |

**Table 1**

**Agreement tasks for the ablation and nesting experiments.** *The first column denotes the name of the task, the second shows the sentence templates, where* `NP` *is used as an abbreviation of* `Det N`. *The indices a, b mark the subject-verb dependencies in the templates. Note that for* Long- *and* Short-Nested, *we test performance on both the* embedded *verb* $V_b$ *and the* main *verb* $V_a$. *The last column contains an example of a sentence in the corresponding agreement task, along with its English translation. Bold and italic face highlight the dependencies marked by the indices in the templates. For each agreement task, we systematically vary the* number *(or gender, in* NounPP-gender*) of all nouns in the template, resulting in four different conditions (SS, SP, PS and PP) for the number-agreement tasks with two nouns (*NounPP-number*,* NounPP-gender*,* Short-Successive *and* Short-Nested*) and eight different conditions (SSS, SSP, SPS, SPP, PSS, PSP, PPS and PPP) for the number-agreement tasks with three nouns (*Long-Successive *and* Long-Nested*). The examples shown are all SS and SSS conditions. For* NounPP-gender, *singular (S) and plural (P) are replaced by masculine and feminine.*

Schmidhuber, 1997), input and output embedding layers of 650 units and input and output layers of size 50000 (the size of the vocabulary). The weights of the input and output embedding layers are not shared (Press & Wolf, 2016). The last layer of the model is a softmax layer, whose activations sum up to 1 and as such corresponds to a probability distribution over all words in the NLM's vocabulary.

### 2.1.4  Model Training

The weights of a NLM are typically tuned by presenting it with large amounts of data (a *training corpus*) and providing feedback on how well it can predict each next word in the running text. This allows it to adjust their weights to maximize the probabilities of the sentences in the corpus. Our NLM was trained on a sample of the Italian Wikipedia text, containing 80M word tokens and 50K word types. Further details can be found in Gulordava et al. (2018).

### 2.1.5  Model Evaluation

Following Linzen et al. (2016), we computed the model's accuracy for the different NA-tasks by considering whether the model prefers the correct verb or adjective form (for the NounPP-number and NounPP-gender tasks, respectively) given the context. We did so by presenting the preamble of each sentence to the NLM and then comparing the output probabilities assigned to the plural and singular forms of the verb for the NounPP task and the probabilities of the masculine and feminine forms of the adjective for the NounPP-gender task. On each sentence, the model was scored 1 if the probability of the correct verb (or adjective) $p_{correct}$ was higher than that of the wrong one $p_{wrong}$, and else 0. The model's accuracy was then defined as the average of these scores across all sentences in the NA-task.

We repeat the above also for another measure, which unlike accuracy, preserves the degree of difference between the two probabilities for the two verb forms. Specifically, we calculate the $success-probability = \frac{p_{correct}}{(p_{correct}+p_{wrong})}$, which results in values in the range $[0, 1]$, where 0.5 corresponds to chance-level performance.

### 2.1.6 Ablations: single-unit and top-k studies

To identify units that play an important role in encoding number or gender, we conducted a series of ablation experiments. In these ablation tests, we assessed the impact of specific units on model performance by setting their activations to zero and then recomputed the performance of the model on the NounPP-noun and NounPP-gender NA-tasks. We conducted two types of ablation studies: single- and multi-unit ablation studies. The first estimated the impact of each unit on model performance, and the latter further assessed the joint impact of groups of units. In the single-unit studies, we separately ablated each of the recurrent units in the network, resulting in 1300 ablation studies per task. For the multi-unit ablations, to avoid the combinatorial divergence implied by ablating all possible unit pairs, triplets, etc., we selected groups of units based on the results from the single-unit study. Specifically, we ranked all 1300 units based on their impact on model performance on the NA-task when ablated in isolation. Then, each time, we selected the top $k$ units from the resulting order, ablated them, and re-assessed model performance on the NA-task. We conducted the multi-unit ablations for $k = 1, ..., 10$. In what follows, we refer to these two studies as the *single-unit* and *top-k ablation study*, respectively, and to the units ordered based on their impact in the single-unit study as the *top-k units*.

## 2.2 Results

### 2.2.1 Ablation Studies Reveal a Sparse Mechanism for Long-Range Number Agreement

We conducted the top-k ablation study with the Gulordava Italian network and with 19 additional NLMs that we trained on the same corpus and using the same hyperparameters, varying random initializations only. In this study, (1) the number of units jointly ablated was incrementally increased, and (2) units were selected for ablation based on their rank in the single-unit ablation results (Figure A3). That is, units with a larger effect on model performance were selected first (Methods). Figure 2 shows the results for all 20 models. We highlight three main findings: First, a significant difference was found between the ablation effects in the SP and PS conditions (continuous vs. dashed lines, respectively), for both the full and the ablated models ($p-value < 0.001$, for all $k \geq 0$; t-test). This difference was consistent across all 20 models, with better performance on the SP compared to the PS condition. Second, for the PS condition, in all cases, model performance dropped to chance level after the ablation of less than 4 units, and the average performance across models reached chance-level performance at $k = 2$ (black dashed line). The mean sparsity of plural encoding for long-range agreements is therefore $< 0.2\%$. Last, for all models, model performance further decreased below chance level for larger $k$'s. This shows that the ablation of a relatively small number of units caused the models to predict the *opposite* grammatical number (singular) in the majority of the trials. All models have therefore developed a 'default bias' towards singular and a sparse mechanism for number propagation across long-range dependencies, mainly dedicated to plural.

### 2.2.2 *The Dynamics of the Long-Range Units Consistently Emerge Across Models*

To confirm that the top-k units are long-range number units, similar to the ones identified in the English Gulordava network, we visualized their dynamics during the processing of the long-range dependency, by extracting their activations during the processing of all sentences in the NounPP NA-task. Figure 3A describes the resulting average cell-state activations for unit 815 from the Gulordava Italian network. Unit 815 had the largest effect ($k = 1$) in the single-unit ablation study (Figure A1). Cell-state dynamics show that number information is robustly encoded throughout the subject-verb dependency, also in the presence of an attractor (dashed lines). The dynamics indicate that unit 815 encodes both singular and plural number, using 'bi-polar' encoding, with negative and positive cell activations, respectively. This is different from the English NLM, that developed two separate 'uni-polar' long-range units specialized on singular and plural, respectively.

Figure B1 further shows unit dynamics for the top units ($k = 1$) from four other models, for both the forget and input gates, as well as for their cell-state activations. In all four examples, the models converged to a similar mechanism with only two variants:

number can be encoded by either bi- (top panel) and uni-polar (bottom) long-range units. The long-range number-encoding dynamics identified in English (Lakretz et al. 2019; Figure 1) thus consistently emerged also in the Italian models, with the same main characteristics: (1) The input-gate activation rises towards its maximal value after the presentation of the main subject. The input gate then resets to zero until the presentation of the verb, thus inhibiting interference from the following attractor; (2) The forget gate maintains high activation throughout the long-range dependency, which is required for storing and propagating the grammatical number of the subject up to the verb; (3) given the input- and forget-gate dynamics, cell-state activation robustly encodes and carries the grammatical number of the subject across the subject-verb dependency. In total, across all models, 19 out of the 20 models show the above dynamics, 7 of which use a bi-polar encoding scheme.[*]

The long-range number-encoding dynamics are also observed in lower-impact units ($k > 1$ in the ablation ranking), showing similar gate and cell-state dynamics. Figure B2 describes gate and state dynamics for the top three units from Gulordava's network.[*] The second top unit ($k = 2$) is a uni-polar unit (middle panel), showing similar long-range plural encoding, however, less robustly – interference from the article before the embedded attractor affects the hidden state. For the third top unit ($k = 3$), long-range number propagation across nested constructions is not observed (right panel). Similarly, in other models, the occurrence of the long-range number-encoding decreases as a function of $k$, and as we show next, unit contribution to verb prediction also decreases as a function of the rank $k$.

---

[*] Gate and cell-state dynamics for all 20 models are available at the repository of the study: https://github.com/yairlak/Nested_Dependencies_in_Humans_and_NLMs/tree/master/figures/top_k_ablation_study/unit_activations/PS

[*] See repository for all models.

### *2.2.3 Long-Range Number Units show a Diminishing Impact on Verb Prediction as a Function of their Rank*

We extracted the efferent weights of the top *k* units, which project onto the output layer. Figure C1A shows an example for the efferent weights of unit 815 from the Gulordava Italian network. It shows that the efferent weights differentially affect unit activations in the output layer, depending on whether they represent singular or plural forms of the verb, which is consistent with its role as a number unit.

We next tested whether the top *k* units from the Gulordava Italian network have a similar or varying impact on verb prediction at the output layer. For fair comparison, we first multiplied each efferent weight by the mean activity of the unit, computed one step before verb prediction (recall that unit activity at the output layer is a function of this product). We refer to this product as the *effective efferent weight*. Figure C2A shows the effective efferent weight for the top ten units identified for the PS condition, from the Gulordava Italian network. In accordance with the top-k ablation results, only the four first units showed strong separation between the two sets of weights, to singular and plural units in the output layer. Importantly, the mean effective efferent weight decreased as a function of the rank *k*, suggesting that the impact on verb prediction diminishes as a function of *k*.[*]

Finally, we repeated the above analysis for all models. Figure C2B shows the mean efferent weight across models as a function of *k*. A monotonic decrease of the mean

---

[*] In a related analysis, we visualized the input and output word embeddings of all target words used in our stimuli. Figures C3 shows these word embeddings represented along two of their principal components (PCs), for the Gulordava's Italian network. Panels A&B show that both number and gender information are encoded at the output word embeddings, separating between the target words of the GA and NA tasks (adjectives and verbs, respectively) with respect to gender and number. Panels C&D show that number and gender information of the main subject of the sentence is separable already at the input word embeddings, although at higher PCs. This suggests that singular and plural forms of the subject, and its masculine and feminine forms, are encoded already at this early processing stage of the network. In the next stage, at the recurrent layers, the long-range mechanism propagate this information across long-range agreements, through possible intervening attractors, as was shown above.

efferent weight as a function of $k$ is observed.

### 2.2.4 Ablation Studies Further Reveal a Sparse Mechanism for Long-Range Gender Agreement

We conducted the single-unit and top-k ablation experiments also for gender encoding, using the gender-agreement task (Table 1). Figure A4 shows the results for all 20 models. On average, for both incongruent conditions (MF and FM), mean model performance sharply drops after the ablation of a single unit, reaching near chance-level performance after the ablation of three units (MF:$0.56 \pm 0.04$; FM: $0.52 \pm 0.02$). This shows that similarly to grammatical number, gender information for long-range agreement is also sparsely encoded, consistently across all models. Figure 3B further shows inner-state dynamics of the top unit from the Gulordava's network. As for the NA-task, the long-range gender unit shows robust encoding across the subject-adjective dependency, also in the presence of attractors (dashed lines). Connectivity analysis further confirmed that the efferent weights of the long-range gender unit are clustered with respect to whether they project to masculine or feminine adjective words in the output layer (Figure C1B). Finally, we note that unlike the case of grammatical number, we did not observe below-chance performance in the top-k ablation study, for one of the incongruent conditions. While on average there's a difference between the two incongruent conditions (compare dashed vs continuous black lines), we did not observe a 'default' bias towards one of the values of gender.

### 2.2.5 Short-Range Number Units are also Found in Gulordava's Network

As was shown for the English NLM, long-range number units are not the only number-carrying units in the network. 'Short-Range' number units also encode grammatical number. However, their activation is not robust to intervening attractors. We therefore tested for the presence of short-range number units in the Italian models, focusing on Gulordava's network. We found 10 more number units, which encode grammatical number in separate values, and whose efferent weights are clustered with respect to number (Figure C1A).

In sum, these results replicate and extend previous findings to another language and another grammatical feature. Similarly to English, only a few long-range number units emerged in the Italian NLMs during training. The NLMs have developed a similar encoding scheme for both grammatical number and gender, independently. Taken together, this shows that sparse long-range *grammatical-feature* units consistently emerge in NLMs.

## 3     Processing of Two Subject-Verb Dependencies in the Italian NLM and in Italian Native Speakers

The core mechanism to carry agreement information across long-range dependencies in NLMs involves a very small number of long-range units. This mechanism can robustly process sentences having a single long-range dependency. However, since the long-range agreement mechanism is sparse, and its small number of units show diminishing impact on model performance, the model is likely to have difficulties to process more then a single number feature at a time due to limited resources. Specifically, we ask: Can the NLM process *two* long-range dependencies that are active at once?

Two simultaneously active long-range dependencies occur in various constructions in natural language, such as center-embedded nested dependencies, a prototypical example of recursion. In nested dependencies, once the long-range agreement mechanism is engaged in tracking the main dependency, there may be no more available resources to process the *embedded* agreement. For example, in the sentence "The **boy** that the *farmer* near the fathers *watches* **knows** the daughters", there are two grammatical numbers that need to be carried across a long-range dependency: (1) that of the main subject 'boy', and (2) that of the embedded subject 'farmer'. Once the NLM encounters the main subject, its grammatical number can be stored through the long-range agreement mechanism up to the main verb 'knows'. However, during this period, since the mechanism is already taken up, once the embedded subject 'farmer' is presented to the network, there is no robust way to encode and carry its number up to the embedded verb 'watches'. The NLM is thus predicted to fail to process the embedded dependency in nested structures.

We emphasize two conditions for this predicted failure:

• The two dependencies are *simultaneously* active at some point: if this is not the case, i.e., the dependencies are successive (e.g., "The **boy** near the cars **says** that the *farmer* near the fathers *watches* the daughters"), the long-range mechanism can first complete processing the first dependency, and then reset before encoding the next one.

• Both dependencies are *long-range*: in the case of a short-range dependency nested within a long-range one (e.g., "The **boy** that the *farmer watches* **knows** the daughters"), the embedded short-range dependency can still be processed by the short-range units we described above, although possibly in a less robust way compared to the main dependency.

We next present the experimental design we used to confirm these predictions about the NLM, as well as to test whether they also extend to human subjects, which would suggest that the agreement processing system of the latter bears similarities to the one we uncovered in NLMs.

### 3.1 Experimental Design

To test the hypothesis that the sparsity of the long-range mechanism can lead to a significant processing difficulty at the embedded dependency, we created a two-by-two design that manipulates the above two conditions: (1) whether the two dependencies are *successive* or *nested*, and (2) whether the embedded dependency is *short* or *long* range. Figure 4 describes the resulting four NA-tasks: Short-Successive, Long-Successive, Short-Nested and Long-Nested ('Short' and 'Long' therefore refer to the length of the embedded dependency only).

The successive tasks serve as control. They minimally differ from the nested ones up to the embedded verb, by only a single word ('dice' ('says') in Figure 4). Note also that tasks that have a long embedded dependency have a third noun, which functions as a possible attractor, inside the embedded dependency. We will refer to this most-deeply embedded noun as the '(inner) attractor', although note that the subjects of the main and nested clauses can also act as attractors on each others' verbs.

For each NA-task, we generated various *conditions* by varying the number of the main and embedded subject noun, and that of the attractor. Short-Successive and Short-Nested have each four conditions corresponding to the possible assignments of number to the main and embedded subjects - SS, SP, PS and PP. Similarly, Long-Successive and Long-Nested have eight conditions, based on the possible numbers of the main, embedded subject and attractor - SSS, SSP, SPS, etc. In what follows, by *congruent subjects* we refer to conditions in which the main and embedded subjects share grammatical number (SS, PP, SSS, SSP, PPS and PPP), and by *incongruent subjects* to the rest (SP, PS, SPS, etc.). By *congruent attractor*, we refer to conditions in which the embedded subject and the third noun share grammatical number (SSS, SPP, PSS and PPP), and by *incongruent attractor* to conditions in which they differ (SSP, SPS, PSP, PPS) (Table 1).

Several studies reported markedness effects in agreement, whereby humans make more errors when the grammatical number of the attractor is plural. This effect was reported in several languages and in both comprehension and production (*English*: Bock and Miller (1991); Eberhard (1997); Wagers, Lau, and Phillips (2009); *Italian*: Vigliocco, Butterworth, and Semenza (1995); *Spanish*: Bock, Carreiras, and Meseguer (2012); Lago, Shalom, Sigman, Lau, and Phillips (2015); *French*: Franck et al. (2002); *Russian*: Lorimor, Bock, Zalkind, Sheyman, and Beard (2008)). Since we use error rates as an index of processing difficulties across various conditions, we strove to increase their overall signal. Therefore, we first present all analyses on contrasts with a plural attractor. Results for the full set of conditions, with both singular and plural attractors, are reported in Appendix D. Also, to facilitate the interpretation of the results, we grouped conditions by whether the main and embedded subjects were congruent or not.

Table 2 summarizes our predictions for each task and structure, while ignoring for now, for ease of presentation, differences among specific conditions. For Short- and Long-Successive, no significant processing difficulties are predicted, since the long-range mechanism can encode the main and nested grammatical numbers sequentially. For Short- and Long-Nested, the long-range mechanism is predicted to successfully process the main dependency and therefore no significant difficulties are predicted on the main verb (beyond the relative difficulty to process center-embeddings reported for both humans (Traxler, Morris, & Seely, 2002) and NLMs (Marvin & Linzen, 2018)). In contrast, the embedded dependency in Short-Nested cannot rely on the long-range agreement mechanism, as it is recruited by the main dependency. Consequently, the processing of the embedded dependency can only rely on short-range mechanisms, which might not be as robust as the long-range ones. We are thus agnostic about success in this case. Finally, in Long-Nested, the performance on the embedded verb is predicted to be significantly low, given that the long-range mechanism can process only a single agreement, as described above.

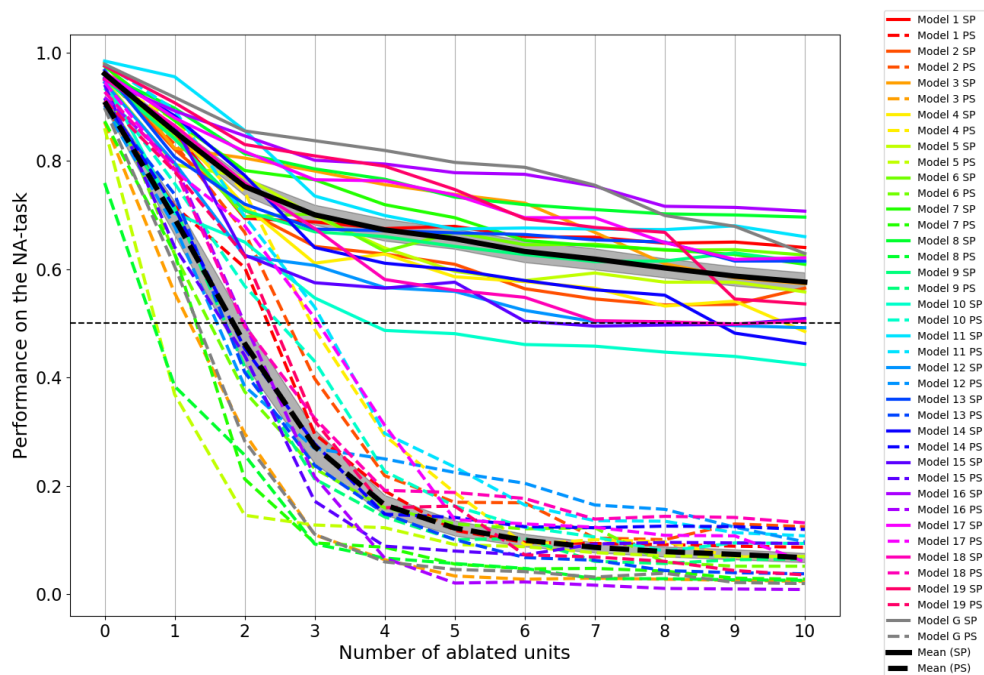

**Figure 2**

***Evidence from ablation about the sparsity of the long-range mechanism and its bias towards singular:*** *To determine the number of recurrent units that participate in the long-range mechanism, we conducted the top-k ablation study (Methods). For all 20 Italian NLMs, model performance on the NA-task was re-evaluated after the ablation of the top k units from the single-unit ablation study ($k = 1, ..10$; $k = 0$ corresponds to the full, non-ablated, model). Model performance is shown separately for the SP (continuous) and PS (dashed lines) conditions. Black lines indicate the average performance across all models, and the shaded grey area represents the corresponding SEM. For the PS condition, mean model performance drops below chance after the ablation of only 2 units (out of 1300 units in the network). The ablation of more units further reduced mean model performance to around 0.1. That is, the removal of a few units caused the models to predict the opposite number (singular) in most trials. This shows that (1) long-range number agreement for plural is sparsely encoded, by less than 0.4% of the recurrent units in the network; and (2) all models developed a 'default bias' towards the singular value.*

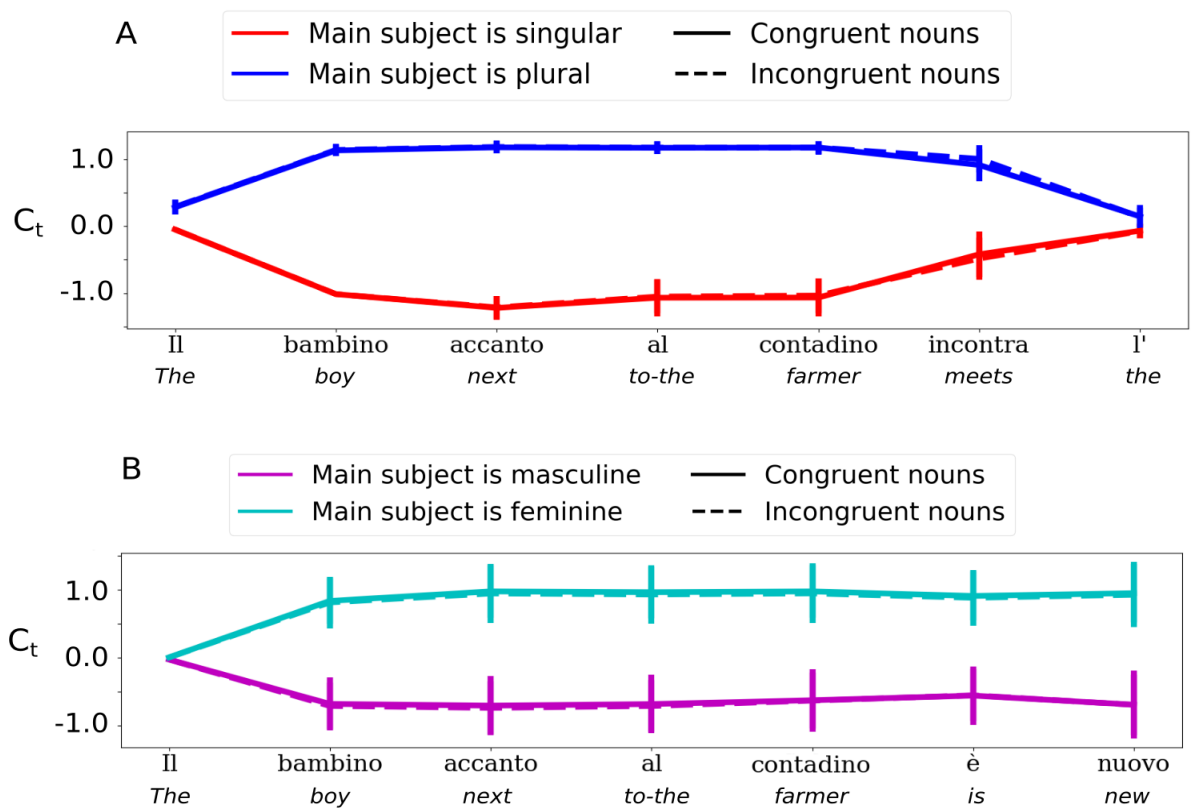

**Figure 3**

***Robust propagation of number (panel A) and gender (panel B) information during the processing of a single long-range dependency across a prepositional phrase.*** *An example sentence for a specific condition (SS in A, masculine-masculine in B) is shown under each diagram, but the plotted cell-state activations are averaged across all stimuli of each condition. (A) Number unit: four conditions are presented, corresponding to whether the main subject of the sentence is singular (red curves) or plural (blue), and to whether the main subject ('bambino') and the attractor ('contadino') have the same (congruent case; continuous lines) or opposite number (incongruent; dashed). Activity of the number unit depends on the grammatical number of the main subject - positive (negative) for plural (singular) value. Activity is stable throughout the subject-verb dependence, also in the presence of an intervening attractor (dashed lines). (B) Gender unit: four conditions corresponding to whether the main subject is masculine (magenta) or feminine (cyan), and to whether the attractor is congruent (continuous) or incongruent (dashed). Activity of the gender unit depends on the gender of the main subject - positive (negative) for feminine (masculine). Activity is stable from the main subject up until the corresponding adjective, also in the presence of an intervening attractor (dashed line).*

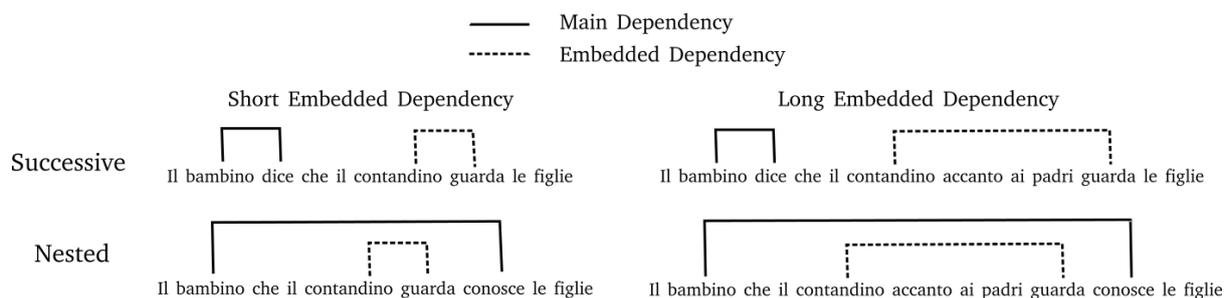

**Figure 4**

***A full-factorial design for two subject-verb dependencies***. *Human subjects and Neural Language Models were presented with sentences from four different syntactic structures, which all have two subject-verb dependencies: a main dependency (continuous lines) and an embedded dependency (dashed). The first factor of the design determines whether the two dependencies are successive (top structures) or nested (bottom), depending on whether the structure has a sentential complement or an object-extracted relative clause, respectively. The second factor determines whether the embedded dependency is short (left side) or long (right). We refer to the four resulting structures as: Short-Successive, Long-Successive, Short-Nested and Short-Long.*

NESTED DEPENDENCIES IN NLMS AND HUMANS 25| **Sentence Type** | **Main Verb** | **Embedded Verb** |
|:---:|:---:|:---:|
| Successive-Short | Good | Good |
| Successive-Long | Good | Good |
| Nested-Short | Good | - |
| Nested-Long | Good | Poor |

**Table 2**

*A summary of the predictions of model performance on successive and nested dependencies based on the sparsity of the long-range mechanism. Cell values represent the degree of predicted performance on the agreement task. Due to possible compensation mechanisms carried by the short-range number units, we make no precise predictions regarding performance on the embedded verb of Nested-Short.*

## 3.2   Methods and Materials for Simulations with the NLM

We used four different Number-Agreement tasks (NA-tasks, bottom, Table 1). All tasks contain two subject-verb dependencies, which differ in terms of whether they are *successive* or *nested* and whether the embedded dependency is *short-* or *long-range*. Two subject-verb dependencies are *successive* when the second subject occurs only after the first subject-verb dependency has been established. Such dependencies occur, for instance, in declarative sentences with a sentential complement, such as "Il **ragazzo dice** che la *donna conosce* il cane" ("The **boy says** that the *woman knows* the dog"). In this example, the subject and verb are directly adjacent in both dependencies. We thus call these dependencies *short-range*, and the above sentence is an example of a sentence that could occur in the *Short-Successive* NA-task. To create sentences that have a successive but *long-range* subject-verb relationship, we increase the distance between the second noun and its corresponding verb by inserting a prepositional phrase in between: "Il **ragazzo dice** che la *donna* accanto al figlio *conosce* il cane' '"("The **boy says** that the *woman* next to the son *knows* the dog"). For nested dependencies, we consider sentences with object-relative clauses, such as "Il **ragazzo** che la *figlia ama* **conosce** il contadino" ("The **boy** who the *daughter loves* **knows** the farmer"). As the nested dependency in this sentence is short range, we call the corresponding number-agreement task *Short-Nested* (note however that the main-clause dependency is long range). For *Long-Nested*, we again use prepositional phrases to increase the distance between the subject and verb of the nested dependency: "Il **ragazzo** che la *figlia* vicino alla donna *ama* **conosce** il contadino" ("The **boy** who the *daughter* near the woman *loves* **knows** the farmer").

For each NA-task, we generated 4096 sentences that were randomly sampled from a fixed lexicon (Table E). The performance of the model on each condition was then evaluated in the same way as described in section 2.1.5.

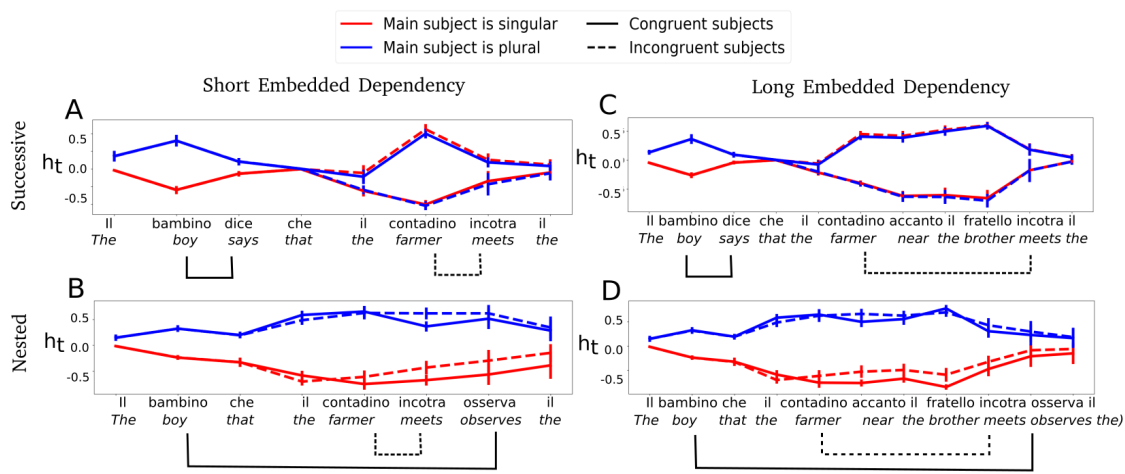

**Figure 5**

***Sequential processing of successive dependencies and robust encoding of the outer dependency of nested constructions by the long-range number unit.*** *Averaged number-unit activity is presented for the four structures in the design: Short-Successive (panel A), Long-Successive (B), Short-Nested (C) and Long-Nested (D). For each structure, results are presented for four conditions, corresponding to whether the main subject of the sentence is singular (red curves) or plural (blue), and to whether the main and embedded subjects have the same grammatical number (congruent; continuous lines) or not (incongruent; dashed lines) (example sentences are for singular/congruent conditions in all cases). (A) Short-Successive: the number unit processes the two successive dependencies sequentially. After the encoding of the first subject ('bambino'), activity return to baseline (around zero) before encoding the embedded subject ('contadino'). (B) Long-Successive: a similar sequential encoding is observed also when the embedded dependency is long-range. (C) Short-Nested: the number unit carries the grammatical number of the main subject ('bambino') up to the main verb ('evita'), also in conditions in which the embedded subject and verb carry opposite grammatical number (dashed lines). (D) Long-Nested: encoding of the grammatical number of the main subject is robust also when the embedded dependency is long-range.*

## 3.3 Results for the Italian NLM

### 3.3.1 The Number Unit Shows Sequential Processing of Two Successive Dependencies and Robust Encoding of the Main Dependency in Nested Constructions

We first simulated NLM dynamics for all NA-tasks and conditions, by presenting all sentences from each condition to the NLM. Figure 5 presents hidden-state dynamics of the top ($K = 1$) long-range number unit during the processing of all conditions from the four NA-tasks. We note that, consistent with results on a single dependency, the grammatical number of the main subject is encoded with the same polarity as found in the NounPP task - positive for plural (blue) and negative for singular (red). Second, in successive NA-tasks, the activation of the number unit returns to baseline after encountering the main verb ('dice'), and then rises back once the embedded subject is encountered. This shows that the number unit can sequentially encode two grammatical numbers in two successive dependencies, also when the two numbers are opposite. In particular, note that in Long-Successive, the grammatical number of the embedded subject is robustly carried across the attractor ('fratello') in the incongruent conditions (dashed lines). Finally, in both Short- and Long-Nested, the grammatical number of the main subject is robustly carried across the main dependency up to the main verb ('osserva'), and in particular, across the embedded subject and verb that carry an opposite number in some conditions (dashed lines). Figure B2 further shows mean unit dynamics of the next two top units ($k = 2, 3$), during the processing of sentences from Long-Nested. The second unit shows long-range number encoding of only the main dependency, similarly to the top unit. However, its functioning is less robust and shows more susceptibility to interference before the embedded attractor. Last, the third top unit does not show long-range number encoding, consistently with the results about the high sparsity of long-range number encoding in the model.

### 3.3.2 NLM Performance on Successive Dependencies

Since no significant processing difficulty is predicted in the successive tasks, we conducted the experiments on the embedded agreement only, which is expected to be the more difficult one. This allowed us to reduce experimental duration in the parallel study with human participants.

Figure 6A presents the resulting NLM error rates for the Gulordava network (Figure D1 shows the full error-rate distribution across all conditions). Figure D3 further provides the mean error-rates across all models. Several effects are observed in the results, for both the Gulordava's network and across models:

- **Near perfect performance on all conditions**: Overall, for both Short- and Long-Successive, the performance of the NLM was found to be almost error free, with slightly more errors in the Long-Successive case ($p < 0.001$)

- **Subject-congruence effect**: We next tested for a *subject-congruence effect*, i.e., whether incongruent subjects elicited more errors compared to congruent ones. We found a significant subject-congruence effect in both Short- and Long-Successive (Short-Nested: $p < 0.001$; Long-Nested: $p < 0.001$). Note that the variance in network performance is very low, which renders this effect significant, although the error rate is near zero in both congruent and incongruent cases.

### 3.3.3 NLM Performance on Nested Dependencies

Figure 6B presents the corresponding error rates in the case of nested dependencies (Figure D1 further provides the full error-rate distribution across all conditions). Several effects are observed in the results:

- **Incongruent conditions elicit more errors**: a significant subject-congruence effect was found on both the embedded and main verbs, in both Short- and Long-Nested (Short-Nested main: $p < 0.001$; Short-Nested embedded: $p < 0.001$; Long-Nested main: $p < 0.001$; Long-Nested embedded: $p < 0.001$). In all cases, incongruent subjects elicited more errors compared to congruent subjects.

- **Processing of embedded verbs is more error-prone**: for both Short- and

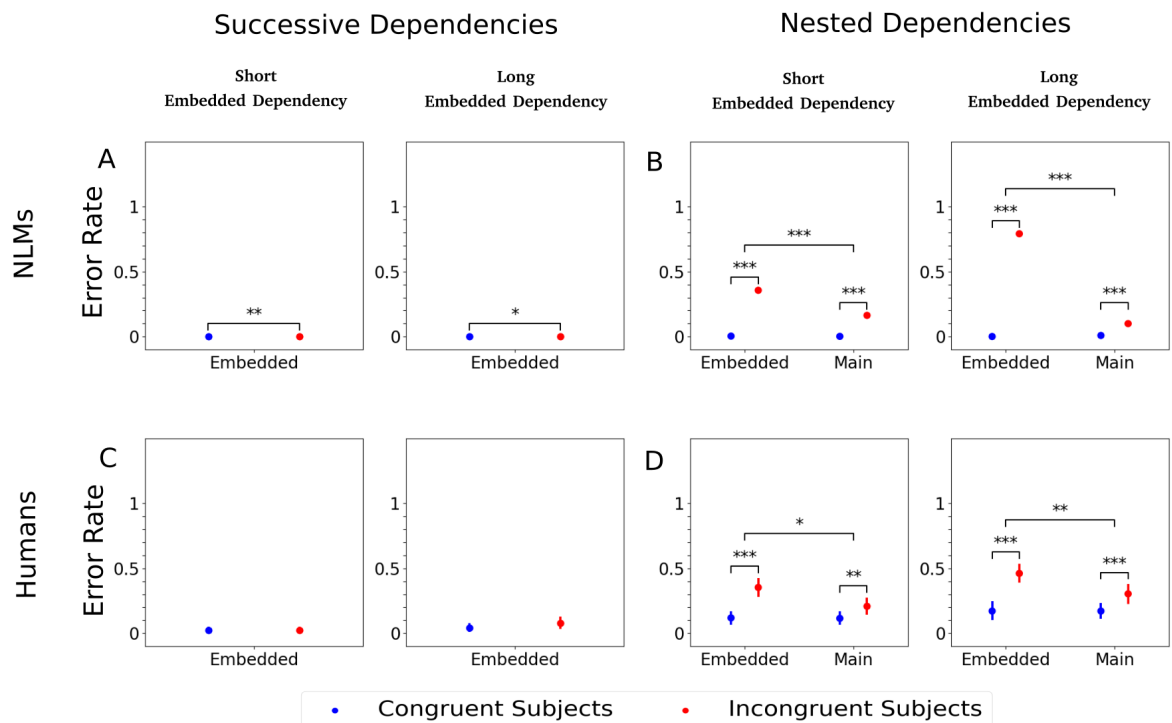

**Figure 6**

**Comparison of error rates** *predicted by the Neural Language Model (top, panels A & B) and observed in human participants (bottom, panels C & D). Blue and red colors correspond to whether the main and embedded subjects agree on number (congruent subjects) or not (incongruent), respectively. Error bars represent the 95% confidence level. Error rates for NLMs were computed based on the output probabilities predicted by the network for the two possible values of grammatical number. For each trial, if the predicted probability of the wrong grammatical number was higher than that of the correct one, the NLM was considered as making an error. Error rates for humans were computed based on the fraction of missed violations in ungrammatical trials.*

Long-Nested, a significant interaction was found between subject-congruence and verb-position (embedded/main), with a larger increase of error rate due to incongruent subjects on the embedded compared to the main verb (Short-Nested: $p < 0.001$; Long-Nested: $p < 0.001$).

- **A longer embedded dependency is more error prone**: for errors made on the embedded verb, we found a significant interaction between subject-congruence and the length of the embedded dependency (Short/Long-Nested), with a larger increase of error rate due to incongruent subjects in Long-Nested ($p < 0.001$).

### 3.3.4 Discussion of NLM results

Focusing on cases with incongruent subjects, the error patterns of the NLM are overall consistent with the predictions summarized in Table 2. First, successive dependencies are relatively easy for the model. This is consistent with the sequential encoding observed in the dynamics of the number unit (Figure 5), which shows a robust encoding of grammatical number of embedded subjects, also in the presence of an attractor. Second, the main dependency in both Short- and Long-Nested was successfully processed (i.e., significantly better than chance level), although with more overall errors compared to successive dependencies. Third, the NLM made an exceptionally large number of errors on the embedded verb in Long-Nested, as predicted by the sparsity of the long-range mechanism. This is prominent in the case of incongruent-subject, since only then the grammatical number of the main subject encoded in the long-range mechanism can interfere. Finally, the NLM made a relatively large number of errors on the embedded verb in Short-Nested but was significantly better than chance level (*error rate* $= 0.5$; $p < 0.001$). The reduced number of errors in Short- compared to Long-Nested is consistent with the findings about a less sophisticated short-range mechanism that can process the embedded dependency, compensating for the momentary recruitment of the long-range mechanism for the main one.

Taken together, this points to a capacity limitation to process nested long-range agreements in NLMs (Lakretz, Dehaene, & King, 2020). We note that a similar capacity limitation was also identified in NLMs trained on artificial data that contained deep nested constructions, with more than two nested long-range dependencies (Lakretz et al., 2021). Specifically, models that were trained on a corpus that contained nested structures up to depth five, failed to generalize to unseen test sentences having a nested structure of greater depth. Therefore, for both natural data and artificial data that contain deep nested structures, NLMs develop a capacity limitation which reflects the corpus on which they were trained.

Finally, we note that in the case of natural language, the capacity limitation observed in the models cannot be simply explained by a limited training corpus if compared to the language input available to humans during language acquisition, as suggested by a reviewer. First, the Wikipedia training corpus contained approximately 80M tokens. Therefore, a single presentation of the corpus to the model involves more tokens than those available to children during their first three years of life (<30M tokens, Hart & Risley, 1995). Second, the vocabulary size of the model (50K token), which was extracted from the training corpus, is of the same order as that of humans (e.g., Brysbaert, Stevens, Mandera, & Keuleers, 2016). Third, long-nested constructions are rare in our training corpus, but that is also the case for natural language (Karlsson, 2007); this makes the human-model comparison fair, ensuring that the obtained results do not depend on different information available to the model vis-à-vis human participants. Last, training on this corpus led to state-of-the-art results on language modelling, and high, human-like, performance on the canonical noun-PP task, on which the models were not explicitly trained.

## 3.4  Methods and Materials for the Behavioral Experiment with Humans

### *3.4.1  Participants*

61 psychology students from the University of Milano-Bicocca (males = 11; Age = 23.3 ± 4.9; Education = 14.6 ± 1.5) took part in the experiment in exchange for course credits. Participants were Italian native speakers and were naive with respect to the experiment purpose. The study was approved by the ethical committee of the Department of Psychology, and ethical treatment was in accordance with the principles stated in the Declaration of Helsinki.

### 3.4.2 *Stimuli*

Stimuli comprised i) acceptable sentences; ii) violation trials, which contained a number violation on the verb of one of the subject-verb agreements; iii) filler sentences, comprising several syntactic and semantic violations. As for the NLM, acceptable sentences were created using a pool of 10 nouns, 19 verbs, 4 prepositions (Table E), for all four main constructions (Table 1, bottom). Starting from the acceptable sentences, number-violation trials were created by replacing either the main or embedded verb by the opposite form of the verb with respect to number. For example, "il **fratello** che lo *studente \*accolgono* **ama** i contadini" ("the **brother** that the *student \*welcome* **loves** the farmers"). Filler trials were based on the same template and they could contain either a semantic or syntactic violation that does not concern number (for further details, see Appendix E). In total, 540 sentences were presented to each participant, randomly sampled from a larger pool. Of these, 180 sentences were acceptable, 180 had a number violation, and 180 were fillers.

### 3.4.3 *Paradigm*

The experiment was conducted in two sessions of 270 trials each, which were performed by participants in different days. Each session lasted around 45 minutes. The two sessions took place at the same time of the day at a maximum temporal distance of two weeks. After receiving information about the experimental procedure, participants were asked to sign a written informed consent.

Stimuli were presented on a 17" computer screen in a light-grey, 30-point Courier New font on a dark background. Sentences were presented using Rapid Serial Visual Presentation (RSVP). Each trial started with a fixation cross appearing at the center of the screen for 600 ms, then single words were presented with SOA=500 ms, 250 ms presentation followed by 250 ms of black screen. At the end of each sentence, a blank screen was presented for 1500 ms, then a response panel appeared, with two labels

'correct' and 'incorrect', on two sides of the screen (in random order each time) for a maximal duration of 1500 ms. A final screen, showing accuracy feedback was presented for 500 ms.

Participants were informed that they would be presented with a list of sentences which could be acceptable or containing a syntactic or semantic violation. They were instructed to press the "M" key of the Italian keyboard as fast as possible once they detected a violation. Sentences were presented up to their end even if a button press occurred before. When the response panel appeared with the two labels ('correct'/'incorrect') at the end of the sentence, participants were instructed to press either the "X" or "M" key for choosing the label from the left or right side of the screen, respectively. During the entire session, participants were asked to keep their left index over "X" and their right index over "M". After each trial, participants received feedback concerning their response: "Bravo!" ("Good!") in case the response was correct, "Peccato..." ("Too bad...") when it was incorrect. At the beginning of each session, participants performed a training block comprising 40 trials. The training section included all stimulus types, which were constructed from a different lexicon than that used for the main experiment.

### 3.4.4 Data and Statistical Analysis

In ungrammatical trials, a violation occurred on either the main or embedded verb. Errors therefore correspond to trials in which a violation was missed. Note that since in ungrammatical trials a violation occurred on only one of the two verbs, the error can be associated with either the main or embedded dependency. In grammatical trials, errors correspond to trials in which participants reported a violation despite its absence. In contrast to ungrammatical trials, in which the violation marks the dependency, in grammatical trials it is not possible to associate an error with one of the two dependencies. Moreover, given the presence of filler trials, the false detection of a violation could be unrelated to grammatical agreement (for example, it could be a false detection of a semantic violation). Agreement errors were therefore estimated from ungrammatical trials only (error rates on grammatical trials are reported in Figure D2).

Statistical analyses were carried out using R (Team et al., 2013). For each hypothesis to be tested, we fitted a mixed-effects logistic regression model (Jaeger, 2008), with participant and item as random factors, using the *lme4* package for linear mixed effects models (D. Bates, Maechler, Bolker, & Walker, 2015). Following Baayen, Davidson, and Bates (2008), we report the results from the model with the maximal random-effects structure that converged for all experiments.

## 3.5 Results for Humans

Section 3.3 showed that the NLM cannot robustly encode two simultaneously active long-range dependencies. The NLM: (1) made more errors on the embedded verb in both Short- and Long-Nested, and (2) had an exceptionally high error rate in the latter case, when the embedded dependency was long range.

Suppose that a similarly sparse mechanism is also used by humans to carry number features through long-distance agreement structures. We derive the following predictions:

- **Prediction 1**: humans will make more agreement errors on the embedded compared to the main dependency in the incongruent conditions of Long-Nested.
- **Prediction 2**: humans will make more errors on the incongruent conditions of the embedded verb when the embedded dependency is long- compared to short-range.

Prediction 1 derives from the robustness of the agreement mechanism in processing the main but not the embedded dependency, as was observed in the NLM performance. The prediction is particularly interesting because, structurally, the main dependency will always be longer-range than the embedded one. Prediction 2 derives from the sparsity of the agreement mechanism and the dramatic failure of the model to process the embedded dependency in Long- but not Short-Nested. Note that we do not make precise predictions regarding Short-Nested, due to possible short-range compensation mechanisms in humans.

In what follows, we describe agreement-error patterns in humans, to be compared to those of the NLM in Section 4.2.

### 3.5.1 Human Performance on Successive Dependencies

Human error rates appear in Figure 6C (Figure D1 further provides the full error-rate distribution across all conditions). Several effects were observed:

- **Relatively low error rate**: humans made relatively few errors on successive constructions, although significantly above zero. Note that, in comparison to the Italian NLM, humans had a higher error baseline, probably due to unrelated factors such as occasional lapses of attention.
- **No subject-congruence effect**: we found no significant subject-congruence effect in Short-Successive, and a marginally significant subject-congruence effect in Long-Successive ($p = 0.06$).

### 3.5.2 Human Performance on Nested Dependencies

Figure 6D presents the resulting human error rates (Figure D1 further provides the full error-rate distribution across all conditions). Several effects are observed in the results:

- **Subject-congruence effect on embedded and main verbs**: for both main and embedded verbs in both Short- and Long-Nested, incongruent subjects elicited more errors–a significant subject-congruence effect was found in all cases (Short-Nested main: $p = 0.003$; Short-Nested embedded: $p < 0.001$; Long-Nested main $p < 0.001$; Long-Nested embedded: $p < 0.001$).
- **Processing of embedded verbs is more error-prone**: for both Short- and Long-Nested, the increase in error rate due to incongruent subjects was larger for embedded compared to main verbs. A significant interaction was found between subject congruence and verb position in both cases (Short-Nested: $p = 0.02$, Long-Nested: $p = 0.008$).
- **A longer embedded dependency is not significantly more error prone**: for embedded verbs, the increase in error rate due to incongruent subjects was comparable when the embedded long-range dependency was compared to the short one. No significant interaction was found between subject-congruence and length of the embedded dependency (Short- vs. Long-Nested).

### 3.5.3 Discussion of Human Results

Overall, as expected, successive dependencies were relatively easy for humans compared to nested ones. The subject-congruence effect was not found, or was marginally significant, which is consistent with sequential processing of grammatical number in NLMs. The results confirm *Prediction 1*–the main dependency in both Short- and Long-Nested was less error prone than the embedded one, as confirmed by a significant interaction between subject-congruence and verb position. However, although humans made more errors on the embedded dependency when it was long range, we did not find a significant interaction between subject-congruence and the length of the embedded dependency. Prediction 2 was therefore not confirmed by the results.

## 4  General Discussion

We investigated how recursive processing of number agreement is performed by Neural Language Models (NLMs), treating them as 'hypothesis generators' for understanding human natural language processing. To this end, we contrasted how NLMs process successive and nested constructions, and tested resulting predictions about human performance.

### 4.1  A Sparse Agreement Mechanism Consistently Emerges in NLMs Across Languages and Grammatical Features

Using agreement tasks with a single subject-predicate dependency, we first replicated in Italian previous findings reported for an English NLM, and extended these findings to another grammatical feature, namely, gender. We found that for both number and gender agreement a sparse mechanism emerged in an Italian NLM during training. These findings suggest that the emergence of a sparse agreement mechanism in the type of NLMs explored in this study is a robust phenomenon across languages and grammatical features [*].

---

[*] Note that in other types of NLMs, e.g., with another type of units (not LSTM), a different mechanism might emerge. The NLM explored in this study is, however, a standard model extensively

The results are moreover consistent with the recent finding that a sparse agreement mechanism emerged in an NLM trained on an artificial language with deep nested structures (i.e., with more than two levels of nesting), suggesting that the sparsity property is not due to the rare occurrence of deep nested structures in natural language (Lakretz et al., 2021).

The top-k ablation study further revealed a default bias towards singular that emerged in all 20 NLMs explored in this study. Such 'default reasoning' in number agreement was previously identified in the Gulordava's English network, using contextual decomposition methods (Jumelet et al., 2019). Interestingly, a 'default' bias to one feature value, known as the markedness effect, was extensively studied also in humans (Bock et al., 2012; Bock & Miller, 1991; Eberhard, 1997; Franck et al., 2002; Lago et al., 2015; Lorimor et al., 2008; Vigliocco et al., 1995; Wagers et al., 2009). The relation between the emergence of default biases in NLMs and humans is an interesting topic for future work. However, we note that, unlike for humans, in our experiments, we did not observe a similar default bias in the case of gender.

The sparsity and specificity of the agreement mechanism suggests that NLMs develop a separate 'module' for at least one specific type of grammatical information. This is not quite a Fodorian module (Fodor, 1983), since the agreement mechanism is not hard-wired and it is not informationally encapsulated. The function of this mechanism can however be characterized independently of the functions of other components in the network, and it can be selectively impaired by neural damage, as was shown in our ablation experiments. This sparse, dedicated representation of syntactic information stands in sharp contrast with the distributed encoding of semantic information, which NLMs typically pack into dense embedding vectors supporting the computation of graded similarity relations (Jurafsky & Martin, 2020; Mikolov, Yih, & Zweig, 2013).

It is an open question whether semantic and syntactic information is encoded and processed jointly or separately in the human brain. At one end of the scale, it was

---

used in applied NLP due to its high performance. The hypotheses emerging from its analysis are therefore compelling for the study of human language processing.

NESTED DEPENDENCIES IN NLMS AND HUMANS                                    39classically claimed that syntactic information is represented and processed in an innate and genetically determined system (e.g., Chomsky, 1984; Fodor, 1983; Pinker, 1994). Both lesion and brain-imaging research have initially supported such a modular view, suggesting that syntactic processing takes place in localized and specialized brain regions such as Broca's area. Early neuropsychological studies showed double dissociations between syntactic and semantic processing. In one direction, aphasic patients were identified with impaired syntactic processing but largely preserved semantic processing (Caramazza & Zurif, 1976). In the other direction, patients were identified with severe semantic impairments but relatively preserved syntactic processing (Breedin & Saffran, 1999; Breedin, Saffran, & Coslett, 1994; Hodges & Patterson, 1996). Further support to this view came from brain-imaging studies, for example, in an influential study, Dapretto and Bookheimer (1999) conducted an fMRI experiment in which subjects had to decide whether two sentences differed in their meaning. In the 'semantic' condition, all pairs of sentences were identical except for one word that was replaced with either a synonym or a different word. In the syntactic condition, sentence pairs were either presented in a different form or a different word order. Brain activations related to the 'syntactic' task were localized to BA 44 in Broca's area, while those for the semantic task were found more anteriorly along the left inferior frontal gyrus, in BA 47. This view became the dominant one in the field, with further support provided from later studies (e.g., Embick, Marantz, Miyashita, O'Neil, & Sakai, 2000; Friederici, Fiebach, Schlesewsky, Bornkessel, & Von Cramon, 2006; Garrard, Carroll, Vinson, & Vigliocco, 2004; Hagoort, 2014; Hashimoto & Sakai, 2002; Pallier, Devauchelle, & Dehaene, 2011; Vigliocco, 2000). However, the original Dapretto et al (1999) study later failed to be replicated (Siegelman, Blank, Mineroff, & Fedorenko, 2019), and in contrast to the modular view, other studies suggested that semantics and syntax are processed in a common distributed system for language processing (e.g., E. Bates & Dick, 2002; E. Bates, MacWhinney, et al., 1989; Dick et al., 2001). This view has recently gained support from brain-imaging studies, providing evidence that semantic and syntactic processing in the language network may not be so easily dissociated from one another (Fedorenko, Blank, Siegelman, & Mineroff, 2020; Mollica et al., 2018). Clearly, there are substantial differences between the human brain and NLMs. However, our findings bring a computational point of view to this debate, suggesting that separating syntactic from semantic processing is computationally advantageous for addressing the language-modeling task, and is spontaneously 'discovered' during the learning process as a way to solve the difficult problem of predicting the next word (see also O'Reilly, 2006; Russin, Jo, & Randall, 2019; Ullman, 2004, for related studies.)

In early debates on the linguistic abilities of neural networks of the 80s and 90s, connectionist theories stressed the parallel nature of neural processing, and the distributed nature of neural representations (e.g., Elman, 1991; Rumelhart et al., 1986a). Addressing recursive processing, Smolensky (1990) formally showed how symbolic, rule-based, algebraic systems (e.g., Fodor & Lepore, 1999; Fodor & Pylyshyn, 1988) can be represented by high-dimensional vectors, and how the generation of syntactic structures can be computed with tensor products between such vectors coding for words, position and role. While these early connectionist theories stressed the importance of the distributed nature of neural representations, our work shows that modern neural networks can develop in certain cases local representations of certain aspects of the input, such as grammatical number or gender. One possible reason for this difference is that modern recurrent neural networks, unlike their early ancestors (Elman, 1990), are equipped with more complex 'innate' mechanisms and structural biases. Specifically, we speculate that gating mechanisms might drive the network towards more local representations of simple features such as grammatical number. Indeed, gating in standard LSTM recurrent networks directly operates on information stored in a *single* unit (see, e.g., Figure 1 in Lakretz et al. (2019)), and thus only indirectly affects information stored in other units of the network. Simple features such as grammatical number would be thus preferably manipulated locally by the gating mechanism. Intriguingly, there are claims about gating mechanisms being present in the brain, both at the micro (Costa, Assael, Shillingford, de Freitas, & Vogels, 2017; Vogels

& Abbott, 2009) and macro (O'Reilly, 2006; O'Reilly & Frank, 2006) levels. An interesting question for future work is thus to directly test the effect of gating on the type of emerging representations (on the local-distributed spectrum) in the network. Finally, we note that other works that studied the English NLM made available by Gulordava et al. (2018), or similar gated RNNs, found that, as with long-distance agreement, the models partially match human behavior, and partially depart from it, also when handling other linguistic phenomena, such as negative polarity item, anaphoric pronoun licensing and filler–gap dependencies (e.g., Futrell, Wilcox, Morita, & Levy, 2018; Marvin & Linzen, 2018; Wilcox et al., 2018). Future research should thus also attempt to explain behavioral differences in terms of the mechanisms underlying these various agreement phenomena."

## 4.2 Processing of Nested Dependencies in NLMs and Humans

We next explored agreement processing in recursive structures that comprise nested subject-verb dependencies. We first confirmed the prediction stemming from the sparsity of the NLM agreement mechanism. The network exhibits exceptional difficulty in processing an embedded long-range dependency within a nested construction. Since NLMs lack a recursive procedure to handle multiple dependencies, once the number units are taken up for the encoding of the outermost dependency, the network fails to process an embedded long-range dependency. In contrast, we found that NLMs achieve relatively good performance on processing *short-range* embedded dependencies, as they can rely on short-range number units outside the core sparse mechanism. The cooperation between the long- and short-range mechanisms in NLMs therefore allows the network to support the processing of a large proportion of agreement constructions in natural language, failing substantially only on relatively uncommon constructions, having two or more nested dependencies that are all long-range.[*]

---

[*] In an analysis of the incidence of embedded clauses, Karlsson (2007) showed that center-embedding constructions are relatively uncommon and multiple nested dependencies are practically absent from spoken language, and rare in written language.

Human results were found to have similarities with the agreement-error patterns of NLMs, but also several important points of discrepancy:

### *4.2.1  Main Similarities*

- **Low error-rates on successive dependencies**: humans and the NLM made a relatively small number of agreement errors on the embedded verb of successive dependencies. For NLMs, this is in accordance with the sequential processing observed in its dynamics (Figure 5). The agreement mechanism resets after the first dependency and is thus available to process the second one. For humans, these findings are in accordance with their relatively good performance on right-branching constructions (e.g., Blaubergs & Braine, 1974; Miller & Isard, 1964)
- **Subject-congruence effect in nested constructions**: in the case of a plural subject attractor, for all verbs, both humans and NLMs made significantly more errors in incongruent cases, in which the main and embedded subjects had opposite grammatical numbers.
- **Higher error-rate on embedded compared to main verbs of nested dependencies**: a positive interaction between verb position and subject congruence was found for both short- and long-range embedded dependencies, suggesting that embedded verbs are more error prone, confirming *Prediction 1*.

### *4.2.2  Main Differences*

- **NLM performance is worse than chance level on the embedded verb of Long-Nested**: the major difference between NLM and human performance (Figures 6 & D1) lies in the behaviour of the NLM with respect to the embedded verb in Long-Nested. The NLM was worse than chance level, meaning that in most trials the network predicts the grammatical number of the embedded verb based on the number of the main subject, which is encoded and carried through by the agreement mechanism. In contrast, human performance is better than chance level (although only marginally so, $p = 0.028$).
- **Prediction 2 was not confirmed in humans**: a strictly related observation is that the

NLM made significantly more errors on the embedded dependency when the dependency was long-range. This was not confirmed in humans, where the interaction between subject-congruence and length of embedded dependency was not significant. Humans as well, however, made more errors in the long-range case.

Bearing in mind that the NLM is trained on raw text data, without being provided with any explicit grammatical knowledge, the points of similarity between the error patterns of humans and the NLM are intriguing. In successive constructions, the sequential processing by the agreement mechanism explains the low error rate of the NLM, similarly to that of humans. In nested constructions, the cooperation between short- and (sparse) long-range mechanisms produces error patterns that are comparable to those observed in humans, with the exception of performance on the embedded verb in the Long-Nested condition. Note in particular that NLMs and humans agree in finding embedded agreement harder than the one in the main clause, despite the fact that the latter is always longer-range.

However, the points of discrepancy raise doubts about whether the agreement mechanism in NLMs could be similar to the one employed by humans. The NLM must have developed this mechanism as a sophisticated solution to the language-modeling task, allowing it to achieve high performance on structures commonly encountered in the data. On such interpretation, the relatively uncommon Long-Nested construction unveils the limitation of the network, pointing to a major difference between it and human subjects. By dissecting the agreement mechanism of the NLM, we could see that it does not support genuine recursive processing. In Short-Nested, the network processes nested dependencies through the collaboration of two *distinct* mechanisms (i.e., short- and long-range). In contrast, a fully recursive mechanism for handling possibly infinite nested constructions, limited only by finite resources, would presumably exhibit self-similarity when processing a subsequent level of the recursive structure.

The issue is whether human performance is, in fact, similarly constrained by nested constructions. One level of nesting, as in object-extracted relative clauses, is known to be relatively difficult to parse by human subjects (e.g., Traxler et al., 2002), and although humans can process two long-range dependencies that are active simultaneously (for example, "The fact that the employee who the manager hired stole office supplies worried the executive"; Gibson, 1998), these constructions are quite demanding, as we also observed in our experiments, and are therefore less common in natural language. Three levels of nesting, such as doubly center-embedded sentences, are known to be nearly impossible to process and to be virtually non-existent in natural language (Karlsson, 2007). Consequently, agreement mechanisms that handle only relatively shallow grammatical dependencies might nonetheless provide a computational solution relevant to human cortical dynamics in some brain regions. If so, the main discrepancy with respect to Long-Nested might turn out to be of a quantitative nature. While NLMs can handle only a single long-range dependency and fail on two, humans can handle two simultaneously active long-range dependencies but would fail on three. Further experiments are required to evaluate such interpretation of the results.

### 4.3 Comparison with Psycholinguistic Theories of Agreement Processing

Several theories have been suggested in the psycholinguistic literature to account for processing difficulties and agreement errors in nested structures. We now discuss our results in light of some of these theories, which may provide complementary high-level explanations compared to that suggested by the neural-network model. It might be worth noting that NLMs have a key advantage compared to existing psycholinguistic theories, as they do not require to assume a grammar or a parsing algorithm. NLMs learn to represent and process underlying structures in natural language by mere training on the language-modeling task, thus minimizing the number of prior assumptions (Lakretz et al., 2020).

#### 4.3.1 Feature Percolation Theories

Early psycholinguistic theories suggested that the proximity between an intervening noun and a verb determines the probability of making an agreement error (Quirk, 1972). This 'linear-distance hypothesis' was later rejected by empirical findings showing that error rates across a prepositional phrase (PP) are higher compared to those across a relative clause, although in the former case the subject is closer to the verb and the syntactic complexity of the preamble is smaller (Bock & Cutting, 1992). Following these findings, Bock and Cutting suggested the 'clause-packaging hypothesis', stating that an attractor within the same clause would generate more interference than one in another structural unit.

More recent studies supported an alternative 'syntactic-distance hypothesis' (Franck et al., 2002; Vigliocco et al., 1995; Vigliocco & Franck, 1999), according to which agreement errors depend on the distance between the head noun and attractor along the syntactic tree, rather than in the linear order of words. According to this view, the grammatical feature of the attractor is assumed to 'percolate' up the syntactic-tree during incremental processing. Such feature percolation can influence the grammatical agreement between the head noun and verb through interference. Feature percolation is assumed to take place incrementally during sentence processing. Therefore, the longer the distance from the attractor to the subject-verb path, the lower the likelihood of interference. The syntactic-distance hypothesis accounts for reduced error rates across relative-clauses compared to PPs, and for additional evidence for which the clause-packaging hypothesis makes inadequate predictions (Franck et al., 2002).

However, percolation theories have difficulties to account for agreement errors on embedded dependencies, such as in Short- and Long-Nested. In these constructions, the attractor with respect to the embedded dependency is the main subject, and thus resides higher on the syntactic tree. As Wagers et al. (2009) note, percolation in such constructions is thus required to happen downwards, whereas percolation theories traditionally assume upward movement through the tree, which could not explain the observed subject-congruence effects. Moreover, syntactic distance between the main and embedded subjects is much greater than that between the subject and the attractor in simple constructions with a prepositional phrase. This predicts lower error rates than those reported for PP constructions. However, our results show higher error rates on the embedded verb compared to previously reported errors on PP constructions in Italian (Vigliocco et al., 1995).

*4.3.2   Memory-based theories*

In sentence comprehension, previous findings reported processing facilitation in ungrammatical sentences due to attraction effects (e.g., Lago et al., 2015; Pearlmutter, Garnsey, & Bock, 1999; Wagers et al., 2009). Self-paced reading paradigms showed that humans process the words that follow the verb faster in the presence of a plural attractor. Importantly, this effect was reported only for ungrammatical sentences. To account for this grammatical asymmetry, previous studies suggested that a cue-based memory retrieval process (Lewis & Vasishth, 2005) is triggered as a repair mechanism following a violation, which, in contrast, would not be triggered in grammatical sentences having no violation. This cue-based memory retrieval process is error prone, and can thus license a wrong verb form in an ungrammatical sentence, explaining the facilitation observed after the verb in ungrammatical sentences only.

Lewis and Vasishth (2005) applied the Adaptive Control of Thought-Rational architecture (ACT-R; Anderson (2013)) to sentence processing, and suggested that, during incremental processing, the transient syntactic structure of the sentence is represented across memory 'chunks' in declarative memory. During sentence processing, each new word triggers a memory retrieval, at the end of which the word is integrated into one of the memory chunks. In the case of verbs, at the end of the retrieval process, the verb will be associated with the appropriate subject stored in memory, ideally, having the same grammatical number. During retrieval, a 'competition' among memory chunks takes place, and the chunk with the greatest number of features matching the verb is most likely to be retrieved. However, erroneous retrievals can occur, due to noise and similarity between memory chunks, in which case a verb carrying the wrong number might be accidentally licensed.

Cue-based retrieval processes were proposed as a repair mechanism, triggered in the case of a violation (Lago et al., 2015; Wagers et al., 2009). For example, for the nested constructions Short- and Long-nested, during the processing of the relative clause, a prediction about the number of the embedded verb is generated. If the embedded verb violates this prediction, as in the case of ungrammatical sentences, a cue-based retrieval is triggered in order to check whether the correct feature was missed. An erroneous retrieval can then license a verb with the wrong number, leading to facilitated reading afterwards. In grammatical sentences, no prediction violation occurs and therefore the repair mechanism will not be triggered, explaining the grammatical asymmetry described above.

In our study, a violation-detection paradigm was used, and therefore a direct comparison with the results from the described self-paced reading experiments and the ACT-R model is not possible. However, we note that processing times on the embedded and main verbs as predicted by the ACT-R model are consistent with our findings. Simulations of sentence processing in the ACT-R model predict greater processing times on embedded compared to main verbs. This increase in processing time is due in part to an extra retrieval cycle associated with retrieving the relative pronoun and attaching a trace to fill the gap in the relative clause structure (Lewis & Vasishth, 2005). An account in terms of cue-based retrieval as a repair mechanism would thus predict more errors on embedded compared to main verbs in nested constructions. However, since processing times cannot be directly mapped onto agreement errors, novel simulation work would be necessary to generate quantitative agreement-error predictions from the ACT-R model, which is beyond the scope of the current study.

A key difference in the error generation process between the two models is that while in the ACT-R based model errors arise during the retrieval process, which occurs after the presentation of the verb, in the NLM agreement errors are estimated one time step before the verb, and are due to a wrong prediction of the next verb. Agreement errors on ungrammatical sentences in the two models are therefore due to different dynamics - erroneous retrievals vs. erroneous predictions. A possible integration of these different dynamics is an interesting topic for future work.

## 5 Conclusion

Our study illustrates how investigating emergent mechanisms in neural-network-based language models can lead to novel explicit hypotheses about linguistic processing in humans, and even to testable predictions about cortical dynamics. The possibility of achieving a mechanistic understanding of natural language processing in modern NLMs can thus inform research in psycho- and neurolinguistics.

Concerning the specific object of our study, we found that the NLM fails to perform genuinely recursive processing of nested constructions. The network develops grammar-sensitive agreement mechanisms for handling constructions up to one degree of nesting only. However, the NLM behaviour matches in part various patterns of human agreement error data, showing remarkable similarity across various sentence constructions. Future research should further probe the nature of the similarities and differences between NLMs and humans, establishing to what extent they are only quantitative in nature, and to what extent they point to a specific adaptation of human neural networks for genuine recursion.

This work was supported by the Bettencourt-Schueller Foundation and an ERC grant, "NeuroSyntax" project, to S.D.

# Appendix A

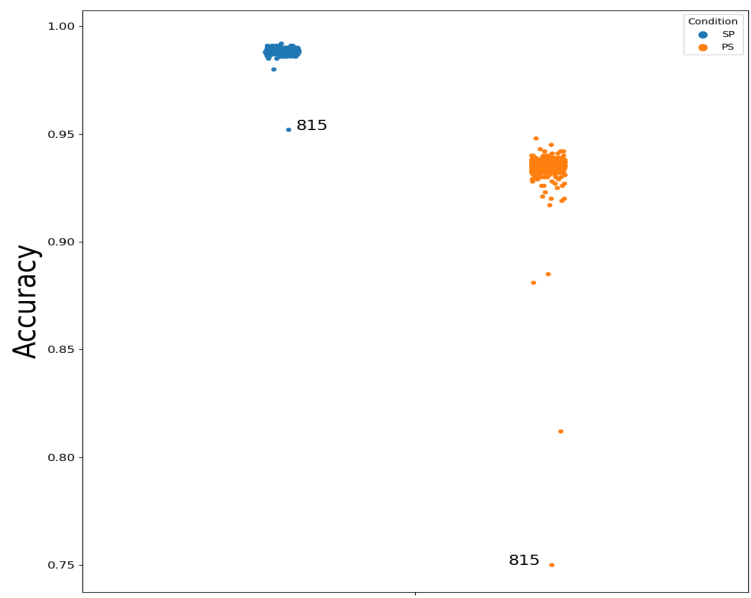

**Figure A1**

***Effect of single-unit ablation on number-agreement performance in the Gulordava Italian network***: *to identify number units, we ablated each time a different unit in the model and tested the ablated model on the number-agreement task. A blue (orange) dot represents accuracy of an ablated model on the SP (PS) condition of the NounPP-number task (Table 1). The ablation of a relatively small number of units led to a significant reduction in model performance ($z-score < -3$; computed based on the distribution of all 1300 single-unit ablation effects). In particular, the ablation of unit 815 from the second layer of the network led to a significant reduction in performance in both incongruent conditions of the NounPP NA-task: for SP, accuracy=0.95 ($z-score = -30.1$; full-model accuracy=0.98); for PS, accuracy=0.75 ($z-score = -27.1$; full-model accuracy=0.94). For the SP condition, unit 815 was followed by one more unit with a significant effect, unit 1119 ($z-score = -6.8$), and for PS, it was followed by three more units - 860, 1119 and 782 ($z-scores = -18.0, -7.9$ and $-7.4$, respectively).*

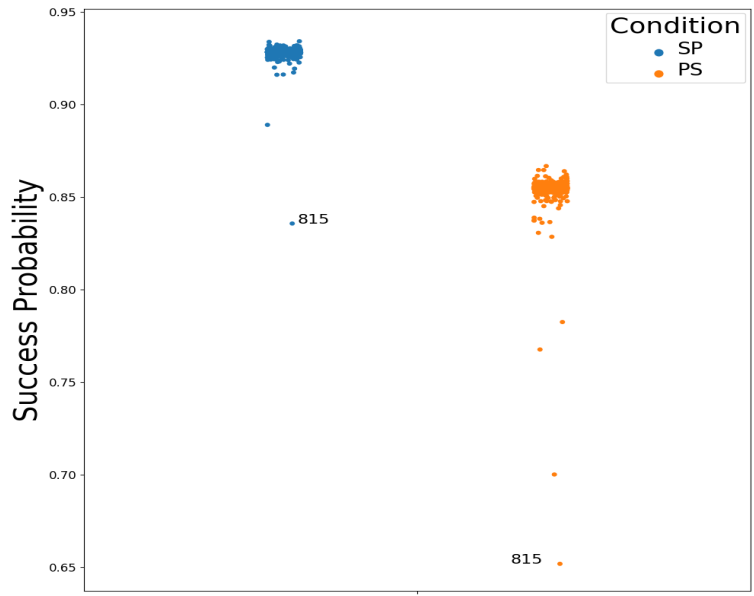

**Figure A2**

***Success probability after single-unit ablations of the model from Gulordava et al. (2018)***: *Same procedure and color scheme as described in the caption of Figure A1. Success probability was defined as $\frac{p_{correct}}{(p_{correct}+p_{wrong})}$ (section 2.1.5), where $0.5$ corresponds to chance-level performance. Similarly to the results from the single-unit ablation study using accuracy as a measure (Figure A1), two units in the SP condition and four units in the PS condition resulted in significant reduction in performance ($z-score < -3$).*

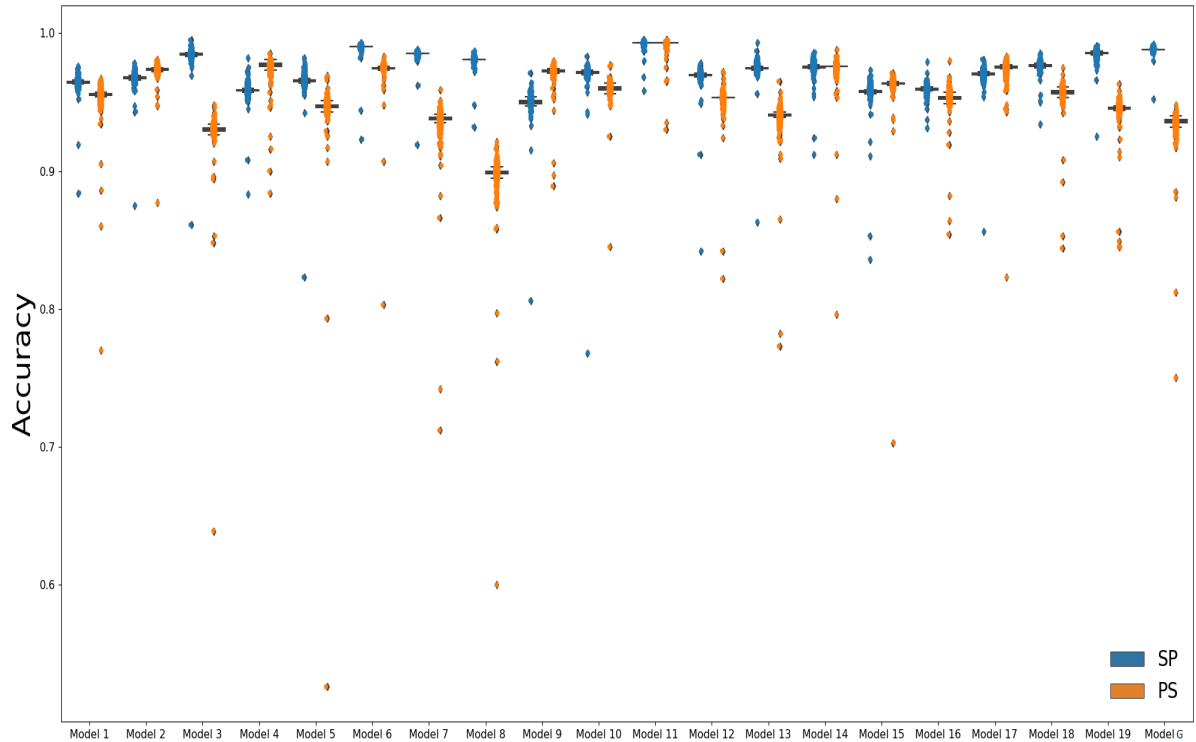

**Figure A3**

***The single-unit ablation study with all 20 models***: *to test the robustness of the sparsity of the long-range mechanism, we trained 19 additional neural language models (NLMs), using the same hyperparameters as those used for training the NLM in Gulordava et al., (2018), Model-G in short. For each NLM, we repeated the single-unit ablation study. Blue and orange dots correspond to accuracy of the ablated models on the SP and PS conditions of the NounPP-number task, respectively (see caption of Figure A1). For all models, for both SP and PS conditions, the ablation of at least one of the units led to a significant reduction in model performance (z-score<-17.3 in all cases). However, unlike for the English Gulordava network, none of the single-unit ablations led to chance-level performance. To further quantify the sparsity of the long-range mechanism, we therefore conducted the top-k ablation study (Figure 2).*

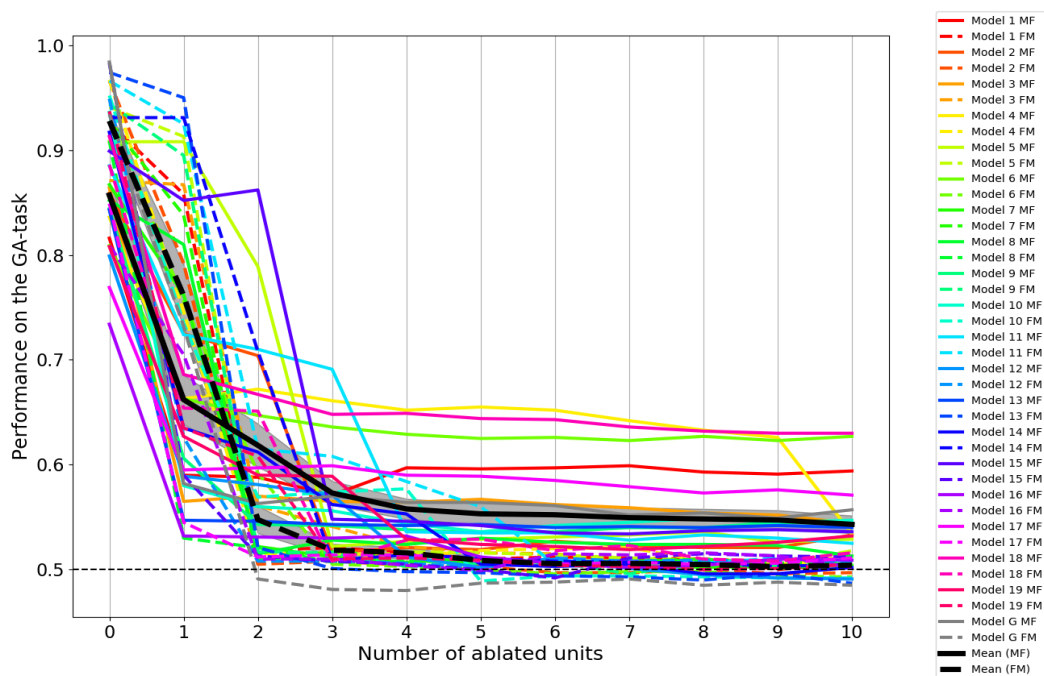

**Figure A4**

***Evidence from ablation about the sparsity of the long-range mechanism for gender:***
To determine the number of recurrent units that participate in the long-range mechanism of gender agreement, we repeated the top-k ablation study for the gender-agreement task (GA-task; Methods). For all 20 Italian NLMs, model performance on the GA-task was re-evaluated after the ablation of the top k units from the single-unit ablation study ($k = 1, ..10$; $k = 0$ corresponds to the full, non-ablated, model). Model performance is shown separately for the MF (continuous) and FM (dashed lines) conditions. Black lines indicate the average performance across all models, and the shaded grey area represents the corresponding SEM. For both incongruent conditions, mean model performance sharply drops after the ablation of a single unit, reaching near chance-level performance (MF: $0.56 \pm 0.04$; FM: $0.52 \pm 0.02$) after the ablation of only 3 units (out of 1300 units in the network). This shows that gender information for long-range agreement is sparsely encoded in the models.

# Appendix B

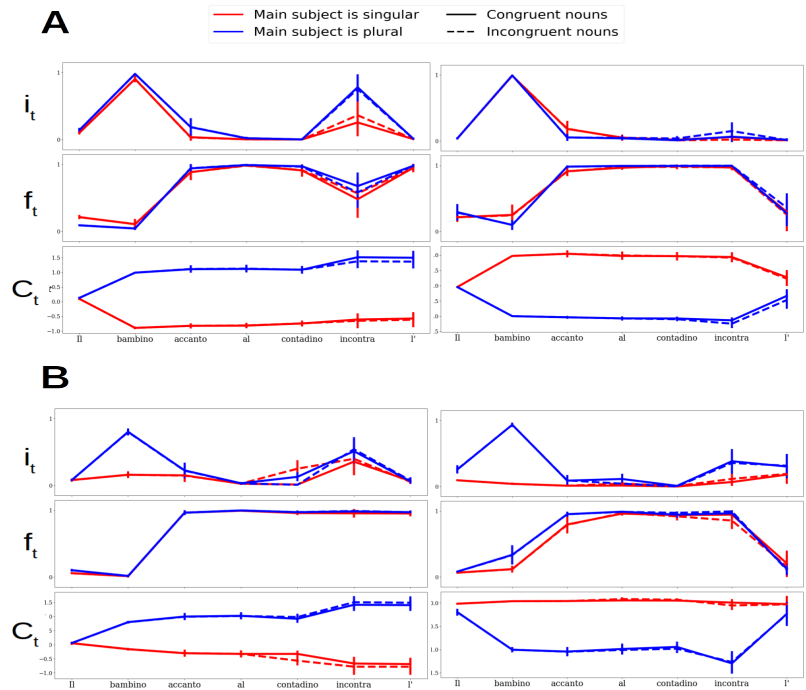

**Figure B1**

**The long-range number-encoding mechanism consistently emerges across models, with only two variants:** *Mean input- ($i_t$), forget-gate ($f_t$) and cell-state ($C_t$) dynamics are shown for the top ($k = 1$) units from the single-unit ablation study, from four models, during the processing of the NounPP NA-task. Error bars represent the standard deviation across trials. The emerging mechanism show similar patterns to those identified in the English model (Lakretz et al., 2019). Specifically, at the main subject, input-gate activity rises towards its maximal value. For 'bi-polar' units, input-gate activity is responsive to both singular and plural subjects (top panels), whereas for 'uni-polar' plural units (bottom), it is selective to plural subjects only. After the subject, forget-gate activity rises towards the maximal value and remains so up until the verb. This is required for continuous memorization of its grammatical number. Finally, input- and forget-gate dynamics lead to sustained cell-state activity from the subject to verb, carrying the grammatical number of the subject across the attractor and up to the verb.*

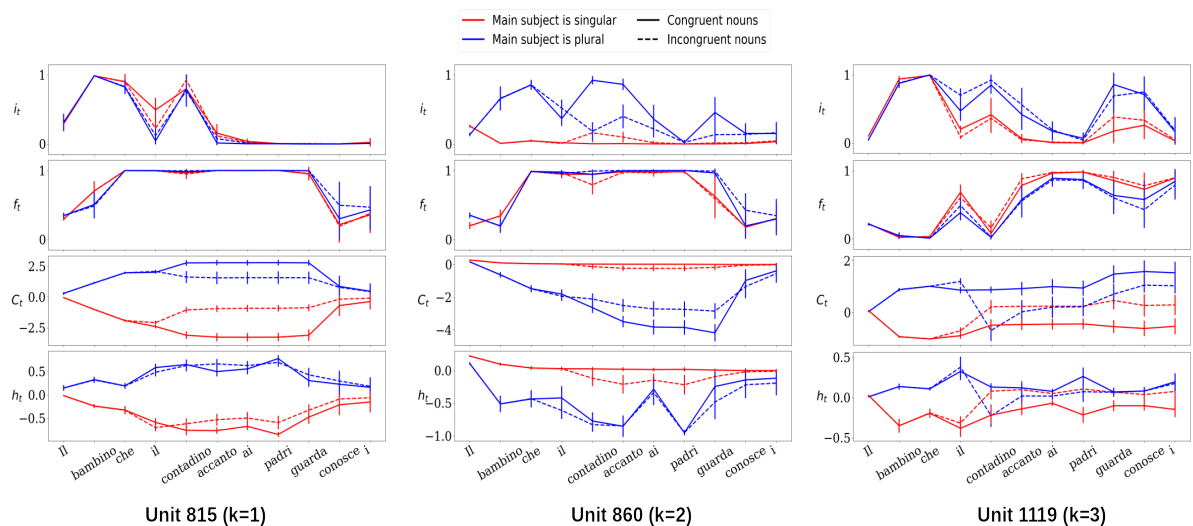

**Figure B2**

***Gate and state dynamics for the first top three units from Gulordava's network:***
*Mean input-gate ($i_t$), forget-gate ($f_t$), cell-state ($C_t$) and hidden-state ($h_t$) activity dynamics are shown for the top three units from the single-unit ablation study ($k = 1, 2, 3$), during the processing of the Long-Nested number-agreement task. Error bars correspond to the standard deviation across trials. The top unit (k=1) shows robust propagation of the grammatical number of the main subject across the main dependency. The second unit (k=2) shows similar patterns but number propagation is less robust before inner attractor ($h_t$ activity at the $7^{th}$ time point). The third top unit ($k = 3$) does not show clear long-range number-encoding patterns.*

# Appendix C

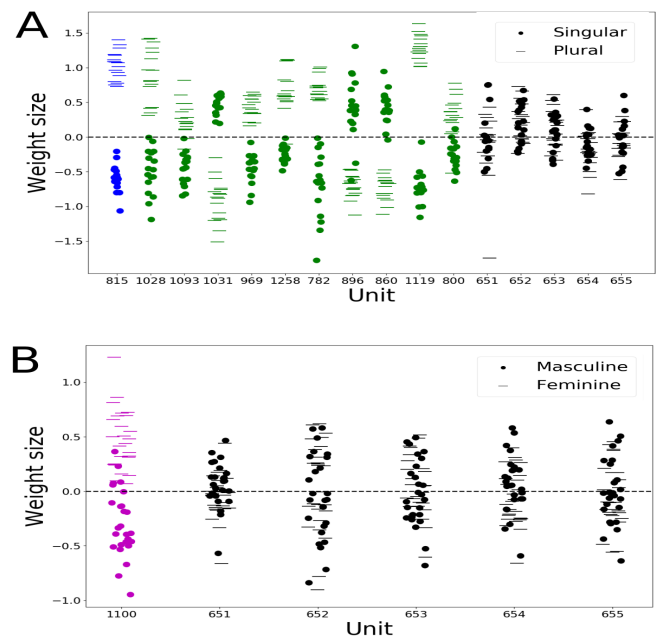

**Figure C1**

***Efferent weights of number and gender units***: *number and gender units were found in the second layer of the network. Units in this layer project onto the output layer, in which each unit corresponds to a word in the lexicon (50,000 in total). We tested whether the efferent weights of the number (gender) units are clustered with respect to grammatical number (gender). A: efferent weights of long-range (815; blue), short-range number units (green) and five arbitrary units (black). Short-range units were identified by detecting units in the network that (1) consistently encode grammatical number in separate values, (2) their activity is sensitive to the last encountered noun, and (3) their output weights are clustered with respect to number, as shown here. All weights project to units in the output layer that correspond to singular and plural forms of verbs. Only weights of number units are clustered with respect to grammatical number. B: efferent weights of the gender unit (1100; magenta) and five arbitrary units. All weights project onto units in the output layer that correspond to masculine and feminine forms of adjectives. Only the weights of the gender unit are clustered with respect to gender value.*

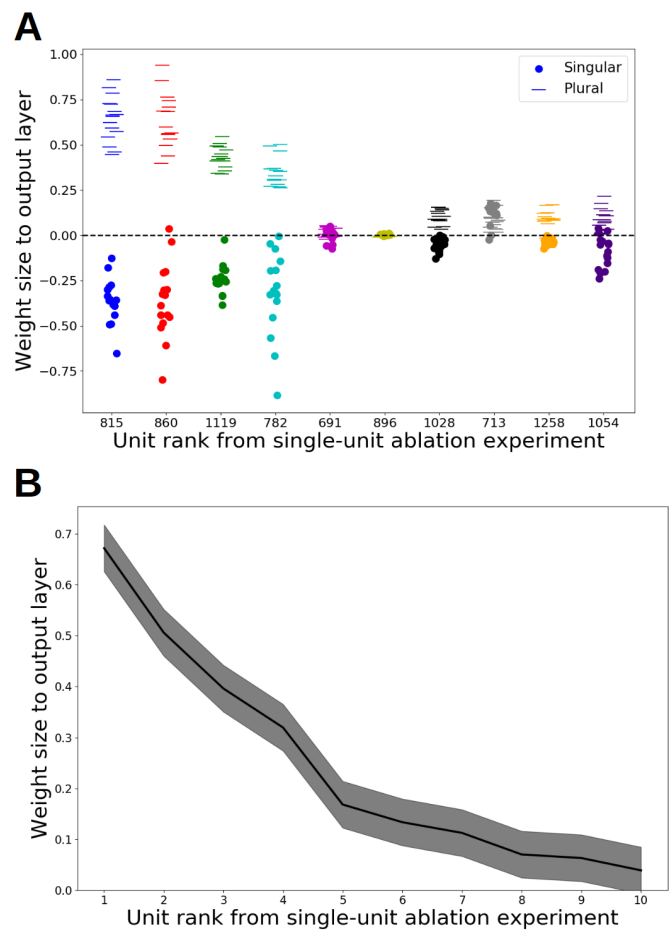

**Figure C2**

***Effective efferent weights of the top 10 units from the single-unit ablation study:*** *A: Effective efferent weights for the 10 top units in the Gulordava Italian network for the PS condition. The effective efferent weight was computed as the product of the efferent weight and the mean unit activity ($h_t$) one time step before verb prediction. The separation between the effective weights to units that correspond to singular and plural forms of the verb decreases as a function of $k$, with largest separation for the top four units. B: Mean effective efferent weights to the output layer, averaged across models. Grey shaded area represents the corresponding SEM. Consistently with the above results from Gulordava's network, a monotonic decrease is observed as a function of $k$.*

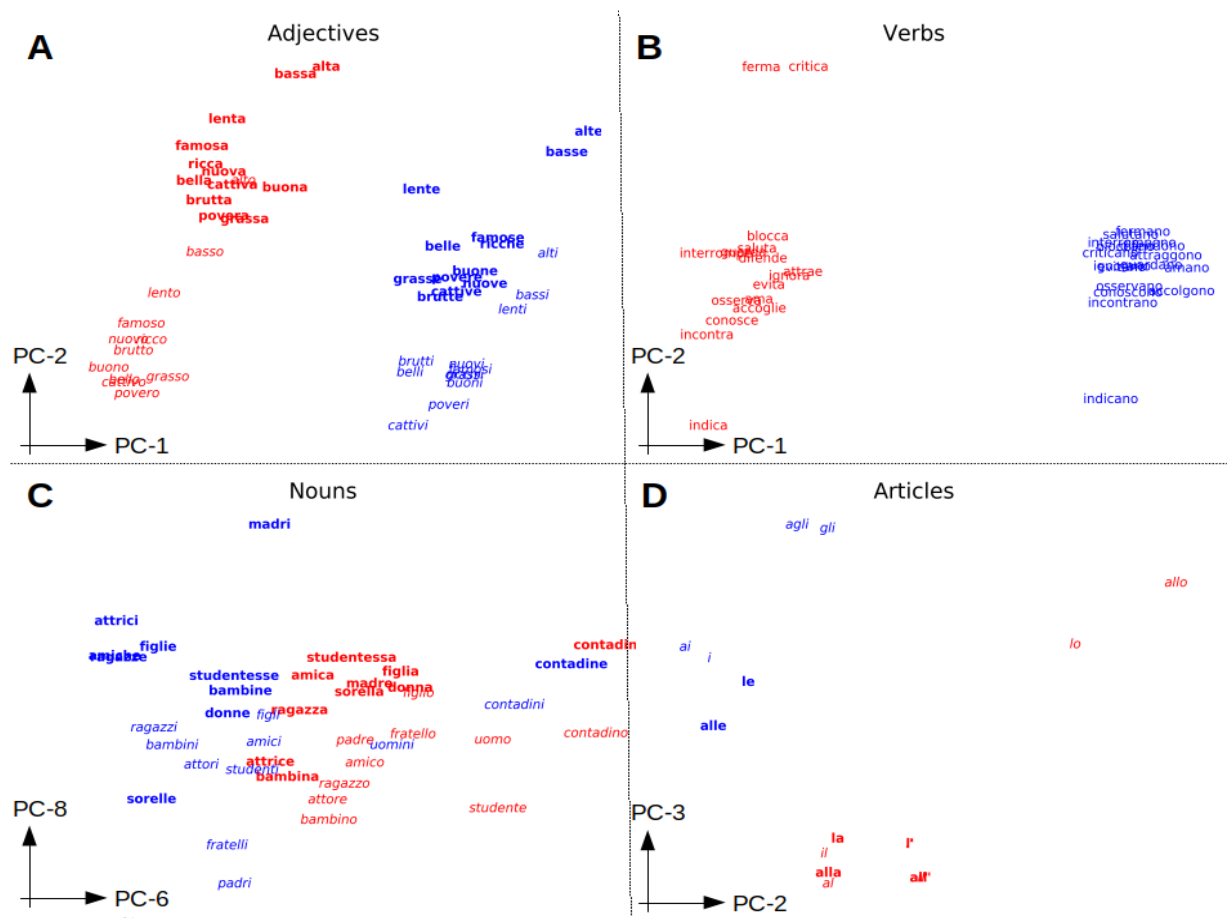

**Figure C3**

***Input and output word embeddings of the Italian Gulordava network:*** *Top - output embeddings: a two-dimensional visualization of the adjectives (panel A) and verbs (panel B) from the number- and gender-agreement tasks are shown. The two leading principal components (PCs) for the 650-dimensional word embeddings are presented. Singular and plural forms of the words are marked with red and blue, respectively. Masculine and feminine adjectives are marked with italic and bold, respectively. For both adjectives and verbs, the first PC separates singular and plural forms of the words. For adjectives, the second PC also separates masculine and feminine, with PC values consistently greater for feminine compared to the corresponding masculine form. Bottom - input embeddings: same color scheme and font styles. For nouns (panel C), the $6^{th}$ and $8^{th}$ PCs are presented; for articles (panel D; including articles in contracted form with a preceding preposition), the $2^{nd}$ and $3^{rd}$ PCs are presented. For nouns, values of the $6^{th}$ PC are consistently greater for the singular compared to the corresponding plural form of the word; and the values of the $8^{th}$ PC are consistently greater for the feminine compared to the corresponding masculine form. For articles, the two PCs separate singular and plural forms of the words.*

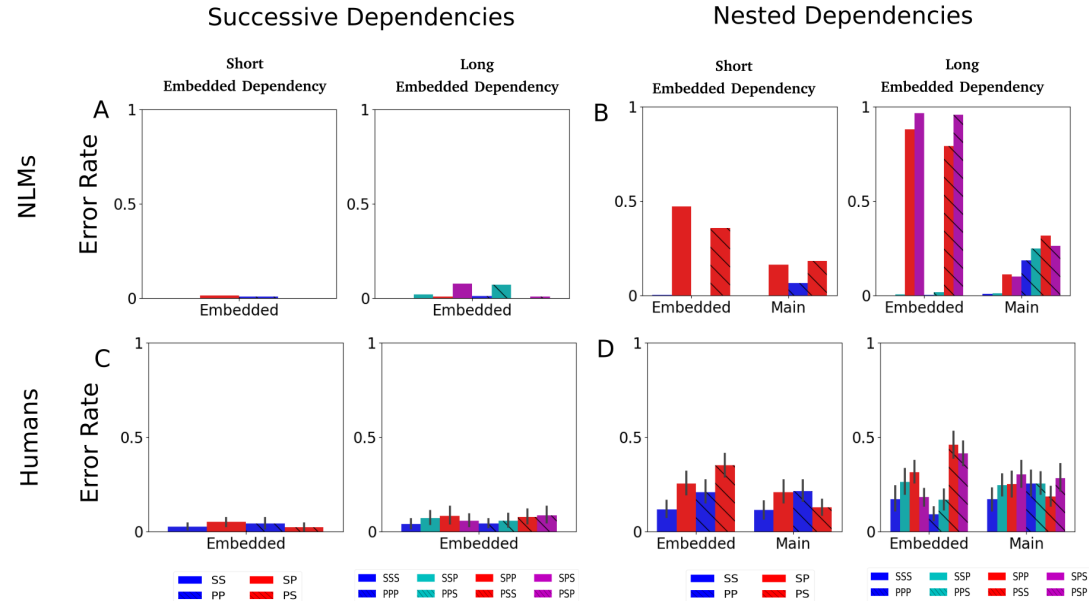

**Figure D1**

**Error rates for all conditions:** *collected from the NLM (panels A & B) and human subjects (C & D). Blue & Cyan (red & magenta) colors correspond to whether the main and embedded subjects agree (don't agree) on number. Secondary colors (cyan or magenta) represent the presence of a nested attractor carrying an opposite grammatical number to that of the embedded subject. Bars with stripes correspond to conditions in which the main subject is plural. Error bars represent the 95% confidence level.*

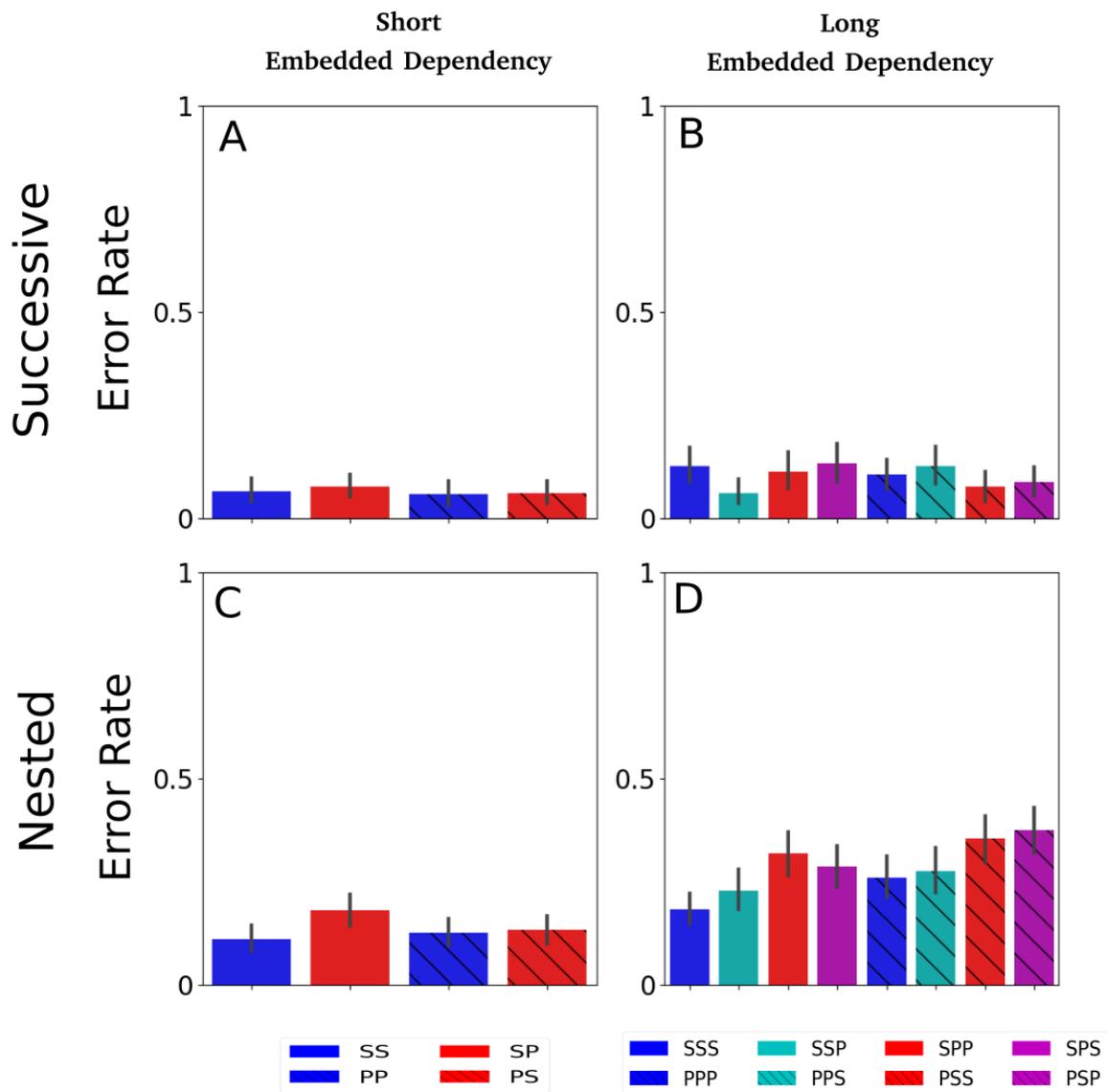

**Figure D2**

***Error rates on grammatical sentences in the behavioral experiment with humans:*** *for the four constructions - Short-Successive (panel A), Long-Successive (B), Short-Nested (C) and Long-Nested (D). Color coding is the same as in Figure D1. Error bars represent the 95% confidence level.*

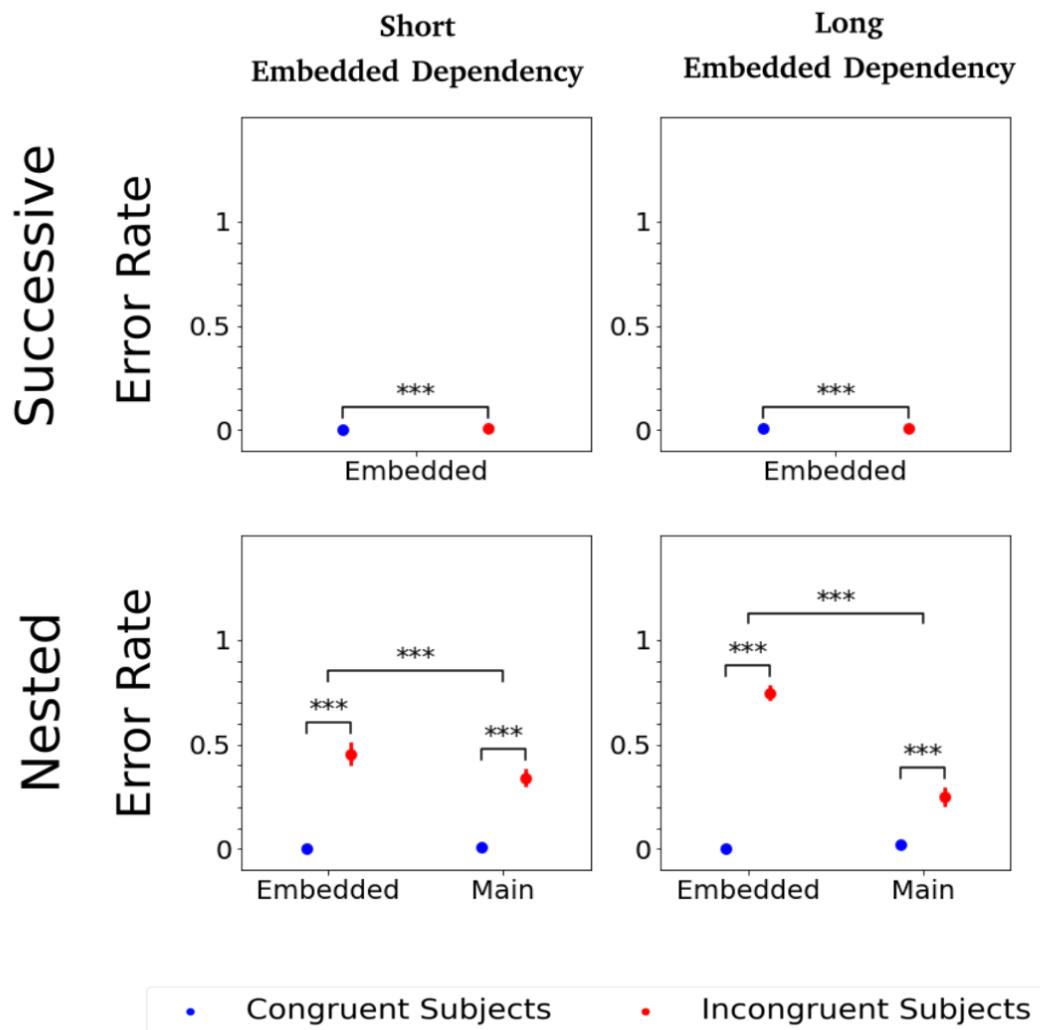

**Figure D3**

**Mean error rates for the 20 models:** *Blue and red colors correspond to whether the main and embedded subjects agree on number (congruent subjects) or not (incongruent), respectively. Error bars represent the 95% confidence level. Error rates were computed based on the output probabilities predicted by the network for the two possible values of grammatical number. Each trial in which the predicted probability of the wrong grammatical number was higher than that of the correct one was counted as an error.*

**Appendix E**

**Materials for the Behavioral Experiment with Humans**

Syntactic violations were generated by either,

i) replacing a verb with a wrong person without changing number, for example: "il **fratello** che lo *studente *accolgo* **ama** i contadini" ("the **brother** that the *student *welcome-1st-pers-sing* **loves** the farmers"); or

ii) replacing a verb with a noun, for example, "il **fratello** che lo *studente *amica* **ama** i contadini" ("the **brother** that the *student *friend* **loves** the farmers"; note that the chosen replacement nouns were not ambiguous with verb forms in Italian); or

iii) replacing a verb with its infinitive form, for example, "il **fratello** che lo *studente *accogliere* **ama** i contadini" ("the **brother** that the *student *to-welcome* **loves** the farmers").

Semantic violations were generated by replacing one of the nouns with either

i) an inappropriate abstract one, for example, "la **\*filosofia dice** che la *figlia ama* la madre" ("**\*philosophy says** that the *daughter loves* the mother"); or

ii) an inanimate noun, for example, "la **\*matita dice** che la *figlia ama* la madre" ("the **\*pencil says** that the *daughter loves* the mother").

To avoid correlation between abstract or inanimate nouns and semantic violations, half of these filler trials were felicitous, for example, "il **padre dice** che la *figlia ama* la \*filosofia" ("the **father says** that the *daughter loves* \*philosophy"), or "il **padre dice** che la *\*matita appartiene* alla figlia" ("the **father says** that the *\*pencil belongs* to the daughter").

| Nouns | masculine | fratello, studente, padre, figlio, ragazzo, bambino, amico, uomo, attore, contadino |
|---|---|---|
| | feminine | sorella, studentessa, madre, figlia, ragazza, bambina, amica, donna, attrice, contadina |
| Verbs | | accogliere, amare, attrarre, bloccare, conoscere, criticare, difendere, evitare, fermare, guardare, ignorare, incontrare, indicare, interrompere, osservare, salutare |
| Matrix verbs | | ricordare, dire, dichiarare, sognare |
| Copula | | essere |
| Prepositions | | vicino a, dietro a, davanti a, accanto a |
| Adjectives | | bello, famoso, brutto, ricco, povero, basso, alto, grasso, cattivo, buono, lento, nuovo |

**Table E1**

*Lexicon used for the agreement tasks. For the nouns, we reported the* singular *forms, in our experiments we use both singular and plural forms. The verb forms reported in the table are the infinitive form, in our experiments, we use the third person singular and plural inflections.*